\documentclass{article}

\usepackage{PRIMEarxiv}

\usepackage[utf8]{inputenc}
\usepackage[T1]{fontenc}
\usepackage{microtype}

\usepackage{fancyhdr}
\usepackage{appendix}

\usepackage{graphicx}
\usepackage{subfig}
\usepackage{siunitx}
\usepackage{multirow}
\usepackage[table,xcdraw]{xcolor} % loads color automatically
\usepackage{booktabs}
\usepackage{colortbl}

\usepackage[backend=bibtex]{biblatex}
\usepackage{hyperref}

\usepackage[acronym]{glossaries}
\glsdisablehyper
\makeglossaries

\newacronym{lic}{LIC}{Learned Image Compression}
\newacronym{fpga}{FPGA}{Field Programmable Gate Array}
\newacronym{clic}{CLIC}{Challenge on Learned Image Compression}
\newacronym{psnr}{PSNR}{Peak Signal to Noise Ratio}
\newacronym{bpp}{BPP}{Bits Per Pixel}
\newacronym{mse}{MSE}{Mean Squared Error}
\newacronym{nlp}{NLP}{Normalized Laplacian Pyramid}
\newacronym{gdn}{GDN}{Generalized Divisive Normalisation}
\newacronym{msssim}{MS-SSIM}{Multi-Scale Structural Similarity Index Measure}
\newacronym{nn}{NN}{Neural Network}
\newacronym{vae}{VAE}{Variational Auto-Encoder}
\newacronym{ae}{AE}{Auto-Encoder}
\newacronym{rd}{RD}{Rate-Distortion}
\newacronym{ml}{ML}{Machine Learning}
\newacronym{kd}{KD}{Knowledge Distillation}
\newacronym{flop}{FLOP}{Floating-point Operation}
\newacronym{iot}{IoT}{Internet of Things}
\newacronym{ai}{AI}{Artificial Intelligence}

\definecolor{Good}{HTML}{32CB00}
\definecolor{Bad}{HTML}{CB0000}

\title{Efficient Learned Image Compression through Knowledge Distillation
\thanks{\textit{\underline{Citation}}: 
\textbf{Authors. Title. Pages.... DOI:000000/11111.}} 
}

\author{
	Fabien Allemand\\
	Télécom SudParis\\
	\texttt{fabien.allemand@telecom-sudparis.eu}\\
	\And
	Attilio Fiandrotti, Sumanta Chaudhuri, Alaa Eddine Mazouz\\
	Télécom Paris\\
	\texttt{\{attilio.fiandrotti, sumanta.chaudhuri, alaa.mazouz\}@telecom-paris.fr}\\
}

\makeatletter
\def\blx@patch@addtocontents{%
  \typeout{Skipping biblatex \string\addtocontents patch (arXiv)}}
\makeatother

\begin{document}
\maketitle

\begin{abstract}
	Learned image compression sits at the intersection of machine learning and image processing. With advances in deep learning, neural network-based compression methods have emerged. In this process, an encoder maps the image to a low-dimensional latent space, which is then quantized, entropy-coded into a binary bitstream, and transmitted to the receiver. At the receiver end, the bitstream is entropy-decoded, and a decoder reconstructs an approximation of the original image. Recent research suggests that these models consistently outperform conventional codecs. However, they require significant processing power, making them unsuitable for real-time use on resource-constrained platforms, which hinders their deployment in mainstream applications. This study aims to reduce the resource requirements of neural networks used for image compression by leveraging knowledge distillation, a training paradigm where smaller neural networks, partially trained on the outputs of larger, more complex models, can achieve better performance than when trained independently. Our work demonstrates that knowledge distillation can be effectively applied to image compression tasks: i) across various architecture sizes, ii) to achieve different image quality/bit rate tradeoffs, and iii) to save processing and energy resources. This approach introduces new settings and hyperparameters, and future research could explore the impact of different teacher models, as well as alternative loss functions. Knowledge distillation could also be extended to transformer-based models. The code is publicly available at: \url{https://github.com/FABallemand/PRIM}
\end{abstract}


%#### Introduction ############################################################
\section{Introduction}
Data compression comes at a cost. There is a limit to how many bits can be removed while encoding data. Beyond this threshold, some data is discarded, potentially degrading the content. This process, known as lossy image compression (as opposed to lossless compression), introduces a tradeoff between reducing file size and maintaining data quality. The GIF format focuses on maximizing compression at the cost of quality, making it suitable for simple graphics but prone to artifacts in complex images like photographs. On the other hand, JPEG or WebP prioritize visual fidelity, compressing intricate images with minimal visible degradation, which often goes unnoticed by the human eye making these algorithms suitable for mainstream use.

\acrfull{lic} sits at the intersection of \acrfull{ml} and image processing. Recent advancements in \acrshort{ml} have led to the emergence of \acrshort{lic}, where \acrfull{nn} are leveraged to perform compression. With an \acrfull{ae} architecture, \acrshort{lic} models can learn how to encode and decode images with different \acrfull{rd} tradeoffs. On the encoder side, the image is projected to a low-dimensional latent space by a convolutional encoder. This representation is quantised, entropy-coded in the form of a binary bitstream and sent to the receiver. At the receiver, the bitstream is entropy-decoded, a convolutional decoder projects such representation back to the pixel domain, recovering an approximate representation of the image. Recent research work present deep \acrshort{nn}s that consistently outperform the most optimised conventional algorithms. However, their high computational demands make them unsuitable for real-time applications on resource-constrained devices, thus preventing their deployment and use for mainstream applications.

Achieving real-time \acrshort{lic} on resource-constrained platforms requires more research in the fields of \acrshort{ml} and hardware specific implementation. This research contributes to this challenge by exploring frugal \acrfull{ai} techniques for \acrshort{lic}. Section \ref{sota} provides an overview of machine learned image compression and state-of-the-art models. \acrfull{kd} is also introduced alongside other related works. We propose a method to apply \acrshort{kd} to \acrshort{lic} in Section \ref{proposed_method}. We disclose our experiment settings in Section \ref{method} and Section \ref{experiments_and_results} presents the results of our experiments.

%#### State-of-the-art ########################################################
\section{State-of-the-art}
\label{sota}

\subsection{Learned Image Compression}
Learned Image Compression (LIC) is a lossy compression method based on machine learning. Using auto-encoder architecture, it is possible to build and train from end-to-end a \acrshort{nn} that achieves balance between image compression efficiency and reconstruction fidelity. \acrshort{lic} consists of three successive steps: projection in a low-dimensional latent space using the encoder, quantisation and lossless entropy coding. Decompression is achieved by applying entropy decoding and projecting the result back into the image space with the decoder \cite{licmedium, licstanford}.

This approach is inspired by transform coding, a signal processing method that consists of three steps: applying an invertible transformation to a signal, quantizing the transformed data to achieve a compact representation, and inverting the transform to recover an approximation of the original signal. This method is used by most deterministic image compression algorithms like JPEG and JPEG-2000.

In 2016, Ballé et al. \cite{balle2016endtoendoptimizationnonlineartransform} propose the first end-to-end optimised image compression framework. Inspired by the field of object and pattern recognition, the framework leverages end-to-end optimisation to achieve better results than previous systems that were built by manually combining a sequence of individually designed and optimized processing stages. Still based on transform coding, the framework consists in transforming an image vector from the signal domain to a code domain vector using a differentiable nonlinear transformation (analysis transform) and applying scalar quantization to achieve the compressed image. The code domain vector can be transformed back to the signal domain thanks to another differentiable nonlinear transformation (synthesis transform). Contrary to traditional methods, the synthesis transform is not necessarily the exact inverse of the analysis transform as the system is jointly optimized over large datasets with the goal of minimising the \acrshort{rd} loss. The rate is measured by the entropy of the quantized vector and the distortion (usually measured using \acrfull{mse} or \acrfull{psnr} in the signal space) is evaluated with either \acrshort{mse} or \acrfull{nlp} after applying a suitable perceptual transform to achieve better approximation of perceptual visual distortion. The authors propose transformations based on \acrfull{gdn} (and its approximate inverse) and to use additive uniform noise at training time to preserve the system differentiability. The first experiments conducted with this framework show substantial improvements in bit rate and perceptual appearance compared to previous linear transform coding techniques.

A few months later, Ballé et al. improve the framework \cite{balle2017endtoendoptimizedimagecompression}. Based on the same three-step transform coding method (linear transformation, quantization, lossless entropy coding), the proposed model uses a nonlinear analysis transformation, a uniform quantizer and a lossless entropy coding. It should be noted that the analysis transformation is inspired by biological visual systems and made of convolutions and nonlinear activation functions (\acrshort{gdn}). By replacing quantization by additive uniform noise at training time (where quantization would have canceled gradients), the model is jointly optimised for \acrshort{rd} performance using bit rate\footnote{The appropriate measure in the context of image compression.} (instead of entropy) and \acrshort{mse}. Although optimizing the model for a measure of perceptual distortion, would have exhibited visually superior performance, \acrshort{mse} was used in order to facilitate comparison with related works (usually trained with \acrshort{mse}) and because there was no reliable perceptual metric for color images. This novel framework yields imperfect but impressive results: details are lost in compression but it does not suffer from artifacts like JPEG and JPEG-2000. It outperforms these codecs at all bit rates, both perceptually and quantitatively according to \acrshort{psnr} and \acrshort{msssim} measures thanks to its ability to progressively reduce image quality.

Driven by the interest of the \acrshort{ml} and image processing communities in machine learning methods for lossy image compression, Ballé et al. extend their end-to-end trainable model for image compression presented in \cite{balle2017endtoendoptimizedimagecompression} with side information \cite{balle2018variationalimagecompressionscale}. Conventional image compression codecs increase their compression performance by sending additional information from the encoder to the decoder. Commonly named side information, it is usually hand designed in these codecs. Using the same formalism as \acrfull{vae}, the authors introduce a more powerful entropy model which acts as a \acrshort{vae} on the latent representation. In other words, it is a prior on the parameters of the entropy model (hyperprior) that is jointly learnt with the main auto-encoder and can be interpreted as side information. This side information is particularly useful as the marginal for an image is likely to be different from the marginal for the ensemble of training images. The additional side information (not seen during training) is valuable for the decoder to reduce mismatch. Once again using the relaxation of the problem (using additive noise instead of quantisation at training time), the authors train different models with and without hyperprior optimised for \acrshort{mse} or \acrshort{msssim} reconstruction loss and for different \acrshort{rd} tradeoffs. \acrshort{psnr} results show that the hyperprior model optimised for \acrshort{mse} consistently outperforms all others \acrshort{lic} methods and performs on par with heavily optimised BGP algorithm. When optimised for \acrshort{msssim}, the hyperprior model is even able to provide better results than state-of-the-art method at all bit rates. The distinction between \acrshort{mse} and \acrshort{msssim} optimised results is relevant as neither have understanding of the semantic meaning of the image, leading to perceptual preferences depending on the image. \acrshort{msssim}, based on human visibility threshold and contrast, attenuates the error in image regions with high contrast, and boosts the error in regions with low contrast yielding good results on images with a lot of textures (like grass) but unsatisfactory results on meaningful high contrast areas like text.

Inspired by success of autoregressive prior in generative models, Ballé et al. update their previous work \cite{minnen2018jointautoregressivehierarchicalpriors}. They generalise hierarchical Gaussian Scale Mixture model to Gaussian Mixture model and add an auto-regressive component. The auto-regressive components captures the context of each pixel, that is to say it allows the model to find spatial dependencies in the image leading to improved image reconstruction. The authors highlight the fact that dimensionality reduction is different from compression. It consists in reducing the entropy of the representation under a prior probability model shared between the sender and the receiver, not only the dimensionality. Experimental results show that the end-to-end optimisation of the model can learn the optimal bottleneck size: if the bottleneck size is large enough, the same latent value is generated and a probability of 1 is assigned for useless channels. This wastes computation but requires no additional entropy. Conversely, small sizes of bottleneck can impact \acrshort{rd} performance. When optimised for \acrshort{msssim}, the proposed model outperforms all conventional and \acrshort{nn} based methods in both \acrshort{psnr} and \acrshort{msssim} (including BPG) in \acrshort{rd} performance as well as visual results.

Early seminal works accounted for a unique latent representation modelled with a fully factored distribution. Since then, much of the research in the field has focused on improving the compression efficiency by refining the entropy model. This basic scheme was then improved by introducing an auxiliary latent space called hyperprior capturing spatial correlation within the image, furthering compression efficiency. \acrshort{lic} has shown the ability to outperform standardised video codecs in compression efficiency, fostering the demand for embedded hardware implementations.

\subsection{Knowledge Distillation}
Applying frugal \acrshort{ai} techniques can help reducing the load of the \acrshort{nn} on the computer. Such methods include pruning, quantisation or \acrfull{kd} \cite{touvron2021trainingdataefficientimagetransformers}. The latter being particularly well suited for \acrshort{lic} as it allows to train a small model to achieve the same performance as a larger one. It should be noted that \acrshort{kd} can be achieved between different architectures, opening the possibility to use a completely different network for decoding \cite{liu2022crossarchitectureknowledgedistillation}.

Originally created to achieve the same results than ensemble of models with a lower computational cost, Hinton et al. \cite{hinton2015distillingknowledgeneuralnetwork} propose \acrlong{kd}. This novel technique consists in transferring the knowledge of a cumbersome model into a single smaller one. In this approach the knowledge of a \acrshort{nn} is not represented by its weights but by the vector to vector mapping it has learned. A large "teacher" model can be trained with unlimited computing power for a long time on large datasets, then a smaller "student" model can learn the mapping of the teacher by using the teacher's predictions as soft targets. This is different than training a smaller model alone as the student model has a higher ability to generalize while requiring fewer and possibly unlabeled data. To compensate the lack of entropy (information) of the soft targets of simple tasks, the authors propose to use a temperature parameter to soften the teacher model output distribution.

In the context of LIC, it is often assumed that the encoding task is performed on a single sender with sufficient resources. The latent representation is then broadcasted and decoded on many receivers with various constraints such as latency and computing resources. Leveraging KD, a smaller and more efficient decoder can be trained while maintaining visual fidelity. Noting that GAN-based LIC frameworks (like state-of-the-art HiFiC) are able to reproduce texture using large general purpose networks, the approach proposed by Helminger et al. \cite{helminger2022microdosingknowledgedistillationgan} overfits a smaller decoder network for every sequence of images that can be sent alongside the latents (more precisely only the blocks responsible of the texture decompression are replaced by a smaller bloc). The smaller decoder is trained using KD on the encoder side. This approach dramatically reduces the decoder model size and the decoding time while providing a great image quality.

More recently, Fu et al. \cite{fu2023fasthighperformancelearnedimage} propose four techniques to improve \acrshort{lic} with resource cautious decoders. They first improve standard \acrshort{lic} by using deformable convolution (convolution with a deformable receptive field) that helps extracting better features and representing objects. Then, a checkerboard context model is used to increase parallelism execution and a three-step \acrshort{kd} method is used to reduce the decoder complexity (train teacher, train student with same architecture of the teacher jointly with the teacher, perform ablation on less relevant blocs of the student decoder and re-train jointly with teacher). Finally, L1 regularisation is introduced to make the latent representation sparser allowing to speed up decoding by only encoding non-zero channels in the encoding and decoding process, which can greatly reduce the encoding and decoding time at the cost of coding performance. The experimental results presented by the authors show better performance than traditional codecs and state-of-the-art \acrshort{lic} methods in both image quality and encoding-decoding time.

\subsection{Other Related Works}
Achieving real-time coding on resource-constrained platforms such as \acrfull{fpga} requires tailored design choices, as demonstrated by state-of-the-art \acrshort{lic} implementations \cite{9745965, 10494759}. Nevertheless, \acrshort{fpga} deployments have generally lagged behind advances in \acrshort{lic}, mainly due to the growing complexity of recent models. Notable exceptions include works such as \cite{mazouz2025lightweight, mazouz2025security, sun2024fpga}.

For example, \cite{minnen2020channelwiseautoregressiveentropymodels} further improves the \acrshort{rd} efficiency by introducing slice-based latent channel conditioning and latent residual prediction with an approach suitable for parallel execution. The \acrshort{rd} efficiency is further boosted in the work of Zou et al. \cite{zou2022devildetailswindowbasedattention} by introducing a Window Attention Module in the \acrshort{ae} architecture and experimenting with a transformer-based architecture in place of the traditional convolutional architecture.

%#### Proposed Method #########################################################
\section{Proposed Knowledge Distillation Method}
\label{proposed_method}
Frugal \acrshort{ai} is an emerging concept that aims to develop models capable of "doing more with less" so frugal models consume less resources while providing great results. \acrshort{kd} is a frugal \acrshort{ai} technique that consists in using the results of a large model to help a smaller model during its training. This model's performance tends towards that of the large model. This research work focuses on applying \acrshort{kd} to \acrshort{lic}. Usually, KD is accomplished by training a small model (student) on the output of a well performing large model (teacher). For classification tasks, the student will have to reproduce the output softmax probabilities of the teacher. This way, the student does not only learn to predict the correct class but also which classes are similar. Following the same reasoning, it makes sense to apply KD to the output of both the encoder and the decoder. The former will assist the student in finding a good latent representation and the latter will help the student to reconstruct input images. The hyper-latent space, corresponding to the bit-stream, can also be subject to \acrshort{kd} in order to achieve higher compression. In the following, we first present our experiment settings after which we analyse the results of our experiments.

%#### Method ##################################################################
\section{Experiments Method}
\label{method}
In this section, we present our methods for consistency throughout the study and for reproducibility purposes. The following parts explain our choice of \acrshort{nn} as well as our training and testing methods.

\subsection{Neural Network}
\begin{figure}
    \centering
    \subfloat[Diagram showing the operational structure of the scale hyperprior compression model. Arrows indicate the flow of data, and boxes represent transformations of the data. Boxes labeled U | Q indicate either addition of uniform noise applied during training (producing vectors labeled with a tilde), or quantization and arithmetic coding/decoding during testing (producing vectors labeled with a hat).]{\includegraphics[width=5cm]{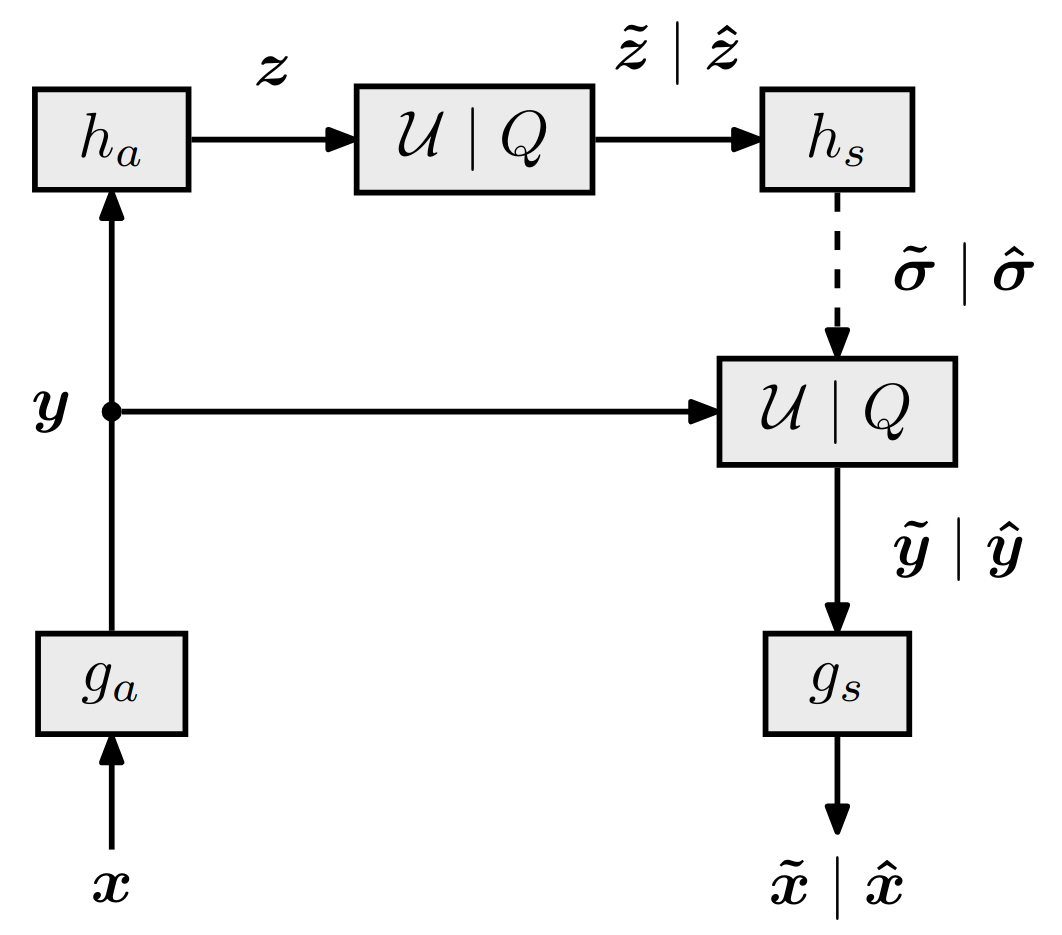} \label{scale_hyperprior_1:a}}
    \qquad
    \subfloat[Detailed architecture of the scale hyperprior compression model. The parameters N in the convolution layers designate the number of channels.]{\includegraphics[width=10cm]{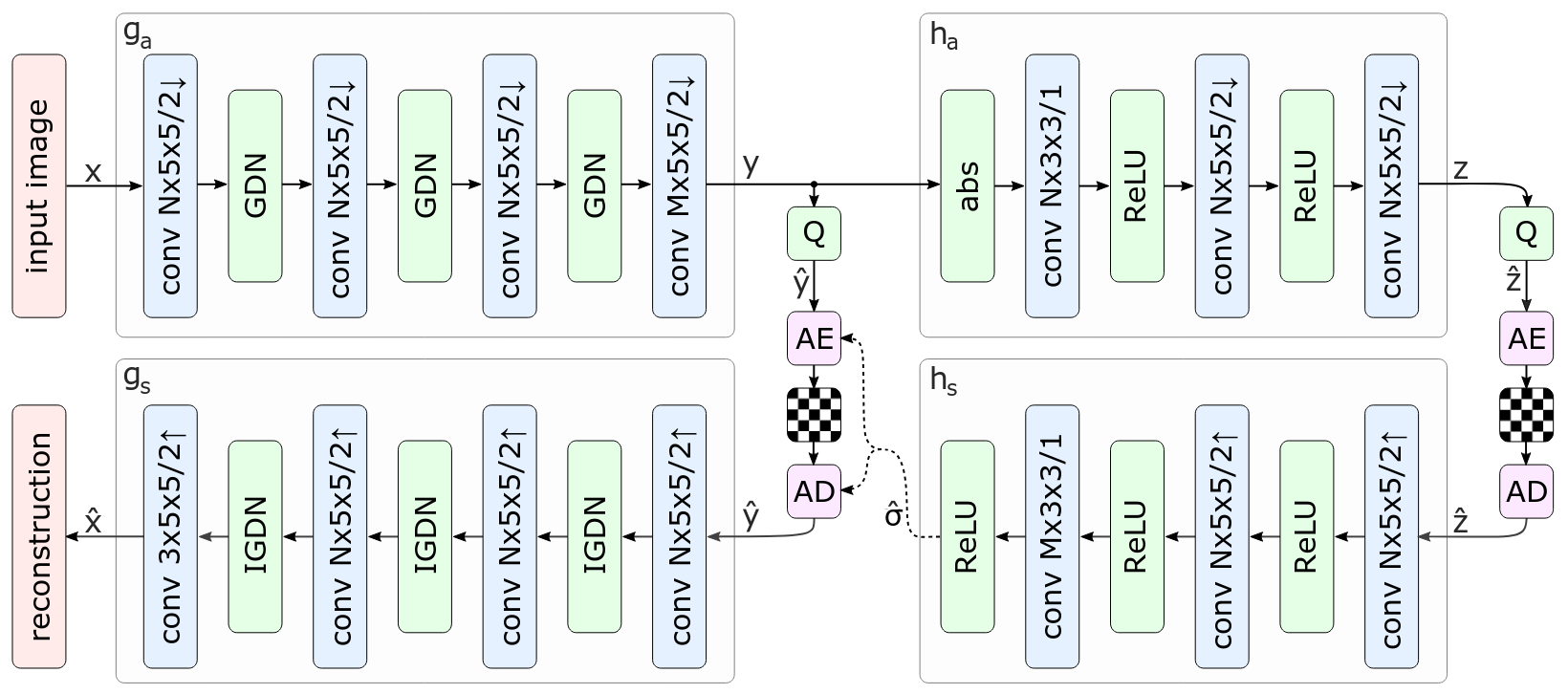} \label{scale_hyperprior_1:b}}
    \caption[Representations of the scale hyperprior model.]{Representations of the scale hyperprior model. (Figures from \cite{balle2018variationalimagecompressionscale})}
    \label{scale_hyperprior_1}
\end{figure}

We selected the scale hyperprior model introduced in 2018 by Ballé et al. \cite{balle2018variationalimagecompressionscale}. This model has the advantage of being straightforward and yet complex enough to be challenging but not impossible to run on resource-constrained computers. Its simple two-stage architecture, shown in Figure \ref{scale_hyperprior_1:a}, makes it a suitable candidate to perform \acrshort{kd} experiments using the latent and hyper-latent spaces.

\subsection{Training Method}
\label{training_method}
\acrshort{kd} requires a teacher model to train smaller student models. We use compressAI pre-trained models as teachers as well as our baseline for comparison.

CompressAI trained their models with two criteria: \acrshort{mse} or \acrshort{msssim}. We keep the models trained with \acrshort{mse}. They were trained using the following loss function: \(L = \lambda\, 255^{2}\, D + R\) with \(D\) and \(R\) respectively the mean distortion and the mean estimated bit rate. The parameter \(\lambda\) allows to adjust the tradeoff between compression and image quality. Higher values give more importance to distortion encouraging the model to produce reconstructed images with high quality at the expense of bit rate. Inversely, lower values of \(\lambda\) imply more compression and more data loss. CompressAI proposes 8 pre-trained models with different values of \(\lambda\) denoted by the argument \textsf{quality}. The correspondence between this argument and the value of \(\lambda\) is summarised in Table \ref{tab_quality_lambda}. It is important to note that the \textsf{quality} argument in compressAI also impacts the network architecture (number of channels and size of the latent space). Pre-trained models with \textsf{quality} lower or equal to 5 have 128 channels and a latent space with 192 dimensions. Other pre-trained models have 192 channels and a larger latent space with 320 components. For easier comparison and because our study focuses on low bit rate image compression, we compare our results to compressAI pre-trained models with \textsf{quality} from 1 to 5.

\begin{table}[t]
    \centering
    \caption{Correspondence between the argument \textsf{quality} in \texttt{CompressAI} and the value of \(\lambda\) when using MSE.}
    \label{tab_quality_lambda}
    \begin{tabular}{ccccccccc}
        \toprule
        \textsf{Quality}             & 1      & 2      & 3      & 4      & 5      & 6      & 7      & 8      \\
        \midrule
        Value of \(\lambda\) for MSE & 0.0018 & 0.0035 & 0.0067 & 0.0130 & 0.0250 & 0.0483 & 0.0932 & 0.1800 \\
        \bottomrule
    \end{tabular}
\end{table}

For fair comparison, we match the training method described in the compressAI documentation \cite{compressai_train} when training the student models. Models are trained between 4 and 5 million steps on 256x256 image patches randomly extracted and cropped from the Vimeo90K dataset. We choose a batch size of 16. The initial learning rate is \(10^{-4}\) and decreases over time (it is divided by 2 when the evaluation loss reaches a plateau). First, we use Equation \ref{loss_1} as the \acrshort{kd} loss function. In later experiments where \acrshort{kd} is extended to the hyper-latent space, we use Equation \ref{loss_2}.

We formalize our distillation objectives as follows. The first loss function combines
feature-level, reconstruction-level, and rate–distortion supervision:

\begin{equation}
    L_{1} = \lambda_{1}\,\mathrm{MSE}(\hat{y}_{student}, \hat{y}_{teacher})
          + \lambda_{2}\,\mathrm{MSE}(\hat{x}_{student}, \hat{x}_{teacher})
          + \lambda_{3}\,\mathrm{RD}(\hat{y}_{student}, \hat{x}_{student}, x),
    \label{loss_1}
\end{equation}

where \(\hat{y}_{student}\) and \(\hat{y}_{teacher}\) denote the latent representations produced by the
student and teacher models, respectively, while \(\hat{x}_{student}\) and \(\hat{x}_{teacher}\) denote
their reconstructed images. The first two terms encourage the student to mimic the
teacher’s latent space and final reconstructions. The last term, \(\mathrm{RD}(\cdot)\),
balances rate–distortion performance by considering both coding efficiency and
reconstruction fidelity. The coefficients \(\lambda_{1},\lambda_{2},\lambda_{3}\) control the
trade-off between distillation fidelity and compression objectives.

\medskip
We extend this formulation by including the hyper-latent space:

\begin{equation}
\begin{split}
    L_{2} = &\;\lambda_{1}\,\mathrm{MSE}(\hat{y}_{student}, \hat{y}_{teacher})
           + \lambda_{2}\,\mathrm{MSE}(\hat{z}_{student}, \hat{z}_{teacher}) \\
           &+ \lambda_{3}\,\mathrm{MSE}(\hat{x}_{student}, \hat{x}_{teacher})
           + \lambda_{4}\,\mathrm{RD}(\hat{y}_{student}, \hat{x}_{student}, x),
\end{split}
\label{loss_2}
\end{equation}

where \(\hat{z}_{student}\) and \(\hat{z}_{teacher}\) denote the hyper-latent representations
used by the entropy model. Compared to Eq.~\eqref{loss_1}, this formulation explicitly
distills not only the main latent features and reconstructions, but also the hyper-latents,
which are crucial for accurate entropy modeling. This richer supervision enables the
student to better approximate the teacher’s probability model and improve compression
efficiency.

\medskip
Finally, the rate–distortion term is defined as

\begin{equation}
    \mathrm{RD}(\hat{y}_{student}, \hat{x}_{student}, x) =
    -\mathrm{E}\!\left[\log_{2}\big(\hat{y}_{student}\big) + \log_{2}\big(\hat{z}_{student}\big)\right]
    + \lambda\,\mathrm{MSE}(\hat{x}_{student}, x).
\end{equation}

Here, \(\mathrm{RD}(\cdot)\) denotes the rate–distortion cost. The first term corresponds
to the expected coding rate, approximated as the negative log-likelihood of the latent
and hyper-latent variables. The second term penalizes the mean squared reconstruction
error between the student’s decoded image \(\hat{x}_{student}\) and the ground-truth
image \(x\), weighted by the Lagrange multiplier \(\lambda\). This trade-off ensures that
the student achieves compact representations without excessively degrading image
quality.

We also tried to use the Kullback-Leibler divergence instead of \acrshort{mse} loss function on the latent space but found that it resulted in slightly higher bit rates. Outcome of this experiment is exposed in Figure \ref{appendix:kd_lic_1_kld}.

\subsection{Testing Method}
In order to compare our results to traditional codecs and other \acrshort{lic} models, different methods can be used. When dealing with lossy compression, it is necessary to assess the visual fidelity of the compressed image compared to the original image as well as other compression techniques outputs. This is done by performing meticulous visual inspection on the entire image for overall evaluation or regions of interest to analyse details. Visual inspection being subjective, it is crossed with quantitative measures. We use standard metrics used in \acrshort{lic} works. Image compression deals with two characteristics: image quality and bit rate. In our study, image quality is measured with \acrshort{psnr}\footnote{We also include \acrfull{msssim} for easier comparison with related work.} and bit rate in \acrfull{bpp}. Together, they define the \acrshort{rd} performance of a compression method. \acrshort{rd} curves are a common tool to compare compression codecs and \acrshort{nn} architectures in \acrshort{lic}. It consists in plotting the average distortion in function of the average bit rate evaluated on a set of images. Using a range of \acrshort{rd} tradeoffs allows to easily evaluate and compare the overall \acrshort{rd} performance of an architecture. We also compute the Bjøntegaard Deltas for both bit rate and image quality, respectively called BD-Rate and BD-\acrshort{psnr}. As explained in \cite{barman2024bjontegaarddeltabdtutorial}, these values characterise bit rate and quality differences between two \acrshort{rd} curves (using \acrshort{psnr} as distortion metric) by measuring the area between the curves.

%#### Experiments and Results #################################################
\section{Experiments and Results}
\label{experiments_and_results}
This section focuses on applying \acrshort{kd} to \acrshort{lic}. To do that, we train a set of student models with \acrshort{kd} to evaluate both their \acrshort{rd} performance and their ability to be deployed on resource-constrained platforms.

\subsection{Rate-Distortion Performance}
By guiding small models during their training with a teacher model, we hope to increase their ability to compress images. We explore two axis: the architecture size and the loss function hyperparameters used to train the students. This section aims at evaluating the \acrshort{rd} performance gain achievable through \acrshort{kd}.

\subsubsection{Architecture Size Tradeoff}
\label{architecture_size_tradeoff}
The objective of \acrshort{kd} is to improve the performance of small models using a large one. However, there is no definition of "a small model" so we start by experimenting with model sizes. Following our training method, we create five new models using \acrshort{kd} and the loss function from Equation \ref{loss_1}. We use \(\lambda = 0.025\) in the \acrshort{rd} part of the loss, the same value used by the teacher model during its training. The five models have different sizes. We modify their architecture by changing the number of channels. The smallest deals with as few as 16 channels while the largest student has 112 (number of channels is indicated by the parameter N in Figure \ref{scale_hyperprior_1:b}). It should be noted that the size of the latent space remains the same across all models, teacher and students.

\begin{figure}
    \centering
    \includegraphics[width=15cm]{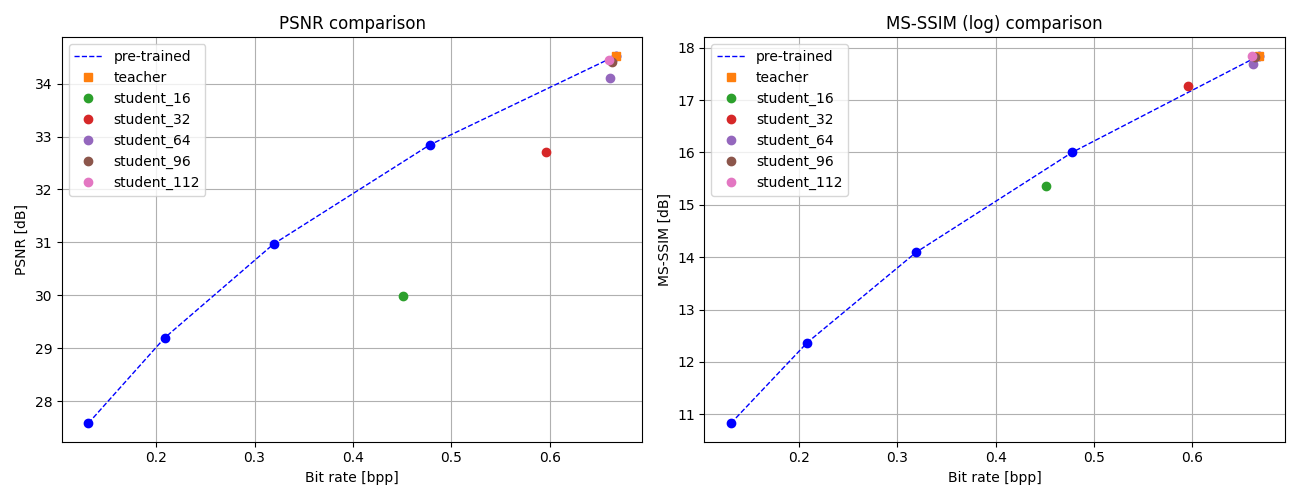}
    \caption[Average \acrshort{rd} curve on the Kodak dataset for students with different number of channels.]{Average \acrshort{rd} curve on the Kodak dataset for students with different number of channels. Despite being trained similarly, all models have different outputs. Models with at least 64 channels performs alike their teacher but 16 and 32 channels are not sufficient to preserve the same image quality.}
    \label{kd_lic_1}
\end{figure}

\begin{figure}
    \centering
    \subfloat[Reconstruction results on image 14 of the Kodak dataset with teacher and student architectures.]{\includegraphics[width=7cm]{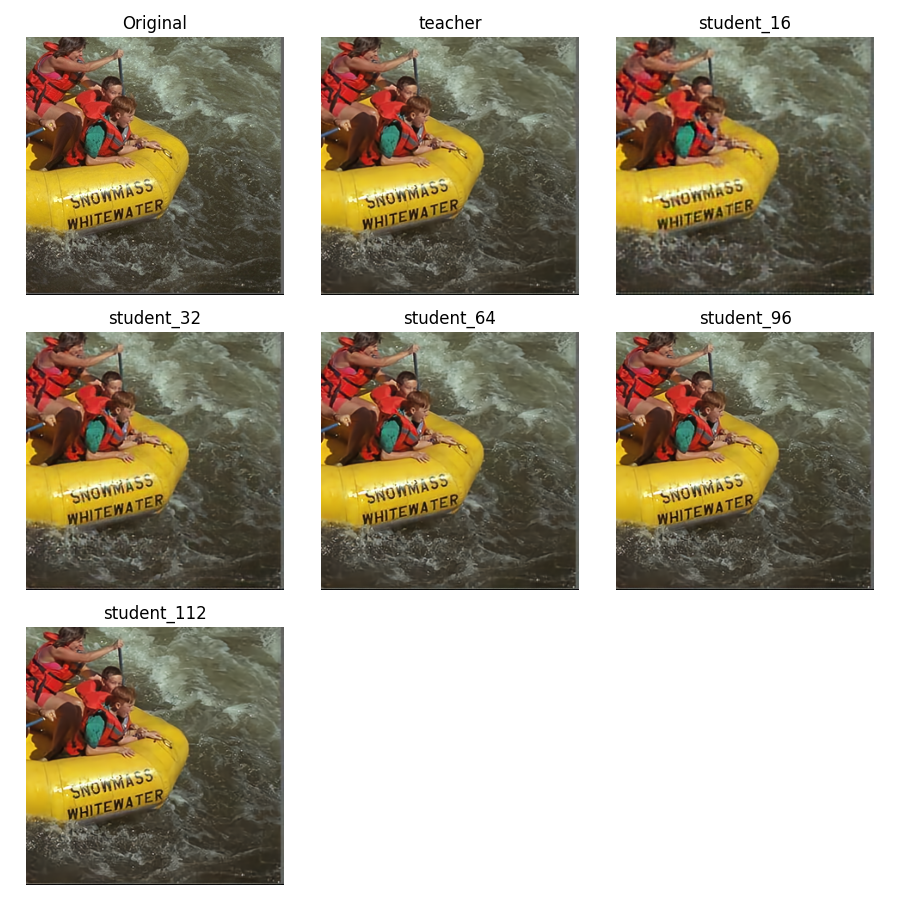} \label{kd_lic_2:a}}
    \qquad
    \subfloat[Average MSE curve on the Kodak dataset for students with different number of channels.]{\includegraphics[width=7cm]{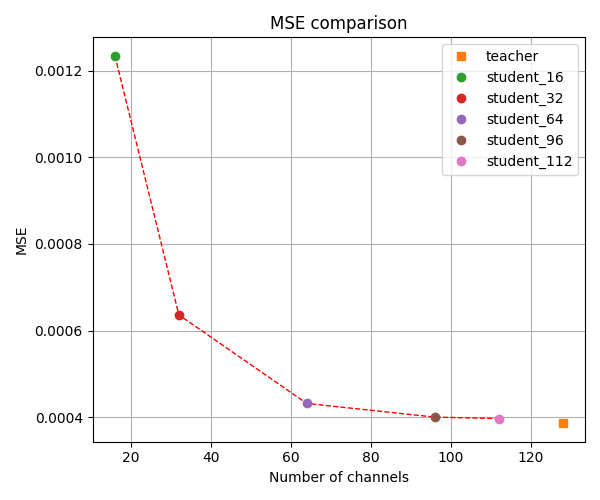} \label{kd_lic_2:b}}
    \caption[Evaluation on the Kodak dataset of the scale hyperprior student models trained for image compression.]{Evaluation on the Kodak dataset of the scale hyperprior student models trained for image compression. Meticulous visual inspection and average \acrshort{mse} of the output prove the importance of the number of channels when limiting compression loss.}
    \label{kd_lic_2}
\end{figure}

Results depicted in Figure \ref{kd_lic_1} follow our intuition. Students with larger number of channels (i.e. 64, 96 and 112) take full advantage of \acrshort{kd} and reach roughly the same \acrshort{bpp} and \acrshort{psnr} as the teacher. Models with 16 and 32 channels cannot reach the same level of image quality. Both of them end up far behind pre-trained models in \acrshort{rd} performance and results in higher \acrshort{mse} error (Figure \ref{kd_lic_2:b}). Even though they produce visually impressive results (Figure \ref{kd_lic_2:a}) for models with so few parameters, they are not relevant candidates when taking into account only \acrshort{rd} performance. The number of channels definitely impact the output quality in extreme scenarios (i.e. very few channels) but can be reduced to some extent without degrading image quality.

\subsubsection{Rate-Distortion and Knowledge Distillation Tradeoffs}
Once again, following our training method, we train five new models using \acrshort{kd} and the loss function from Equation \ref{loss_1}. The teacher model is still taken from the compressAI zoo with \textsf{quality} set to 5. This time, we fix the number of channels across all models (i.e. 64, the smallest number of channels that did not degrade too much performance in previous experiments) but we use different values of \(\lambda\) in the \acrshort{rd} part of the loss. We use the pre-trained models corresponding to the same values of \(\lambda\) as a baseline to compare our results.

\begin{figure}
    \centering
    \includegraphics[width=15cm]{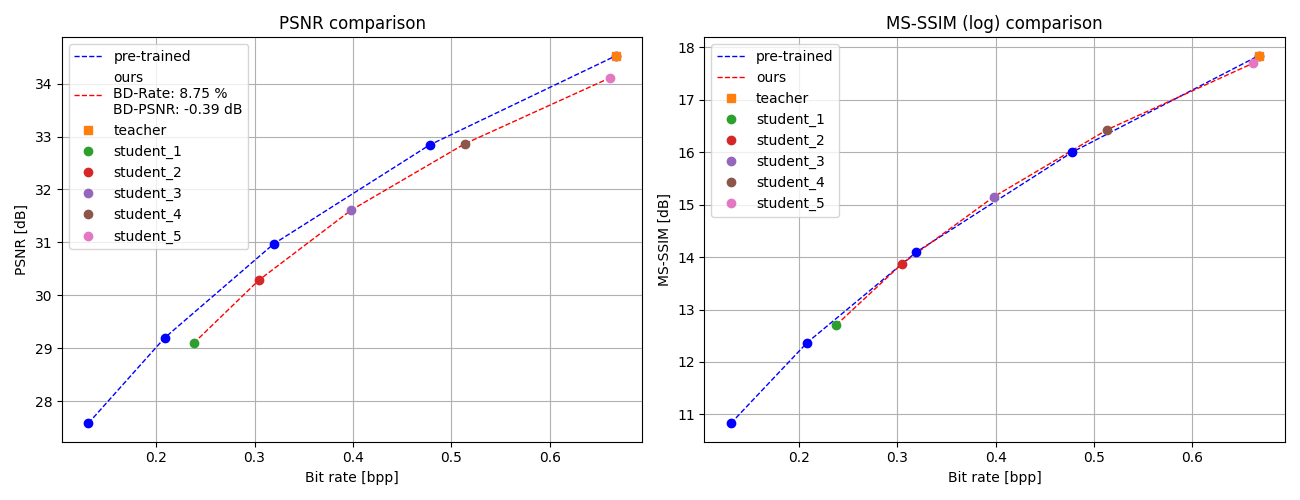}
    \caption[Average \acrshort{rd} curve on the Kodak dataset for students with different \acrshort{rd} tradeoffs.]{Average \acrshort{rd} curve on the Kodak dataset for students with different \acrshort{rd} tradeoffs. Trained by a teacher with an emphasis on image quality, students 1, 2 and 3 achieve better \acrshort{psnr} than their pre-trained counterpart. Students 4 and 5 are held back by the limited number of channels.}
    \label{kd_lic_4}
\end{figure}

\begin{figure}
    \centering
    \includegraphics[width=15cm]{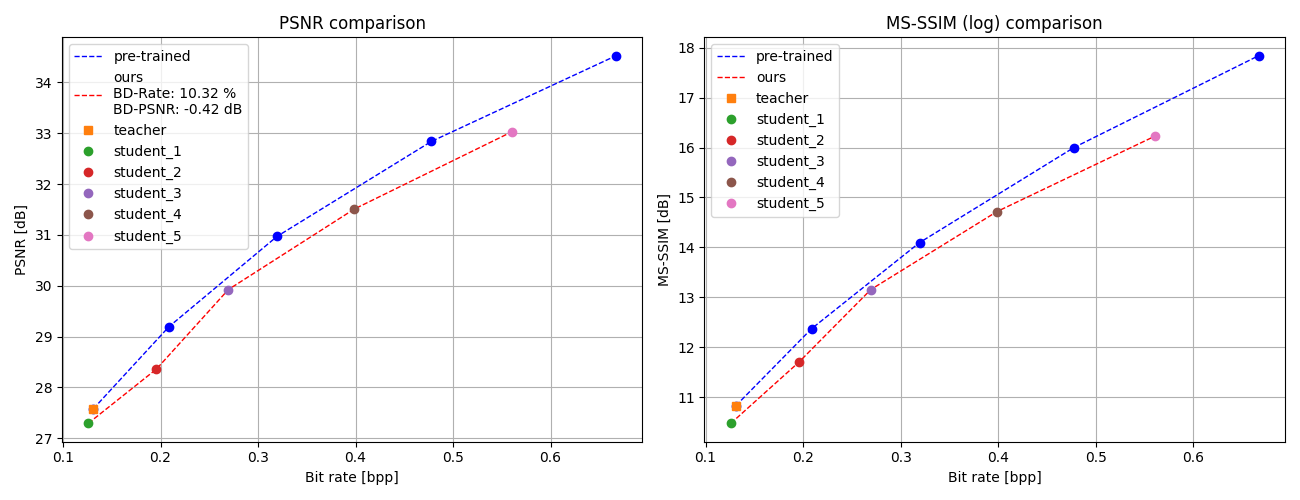}
    \caption[Average \acrshort{rd} curve on the Kodak dataset for students with different \acrshort{rd} tradeoffs and a teacher focusing on minimising bit rate.]{Average \acrshort{rd} curve on the Kodak dataset for students with different \acrshort{rd} tradeoffs and a teacher focusing on minimising bit rate. Student \acrshort{nn}s, influenced by they teacher at training time, have a lower bit rate than their pre-trained counterpart. The gap becomes more noticeable as \(\lambda\) increases. This gain in bit rate implies a loss in image quality across all students.}
    \label{kd_lic_4_bis}
\end{figure}

As a reminder, the pre-trained models in Figure \ref{kd_lic_4} all have 128 channels. This means that our student models all have half as many channels as the pre-trained models do. Still, all students have similar or higher \acrshort{psnr} than their pre-trained counter-part. More precisely, when quality is prioritised (students 4 and 5), \acrshort{kd} is not the ideal solution: the models performs better than what they would have if trained alone but they are restrained by the modest number of channels. \textsf{student\_5} attains the limit of what is feasible with 64 channels: with the same \acrshort{rd} tradeoff as the teacher, it is not able to match the teacher image quality. However, smaller models (student 1, 2 and 3) trained to prioritise bit rate reach better \acrshort{psnr} thanks to the teacher model (trained to favorise image quality) pulling them toward the top of the chart. Another interesting experiment is to use a teacher trained to prioritize bit rate. In that case, we observe that all students are pushed toward the bottom left of the chart (Figure \ref{kd_lic_4_bis}). This compromise could be useful for storage or bandwidth critical operations.

\begin{figure}
    \centering
    \includegraphics[width=15cm]{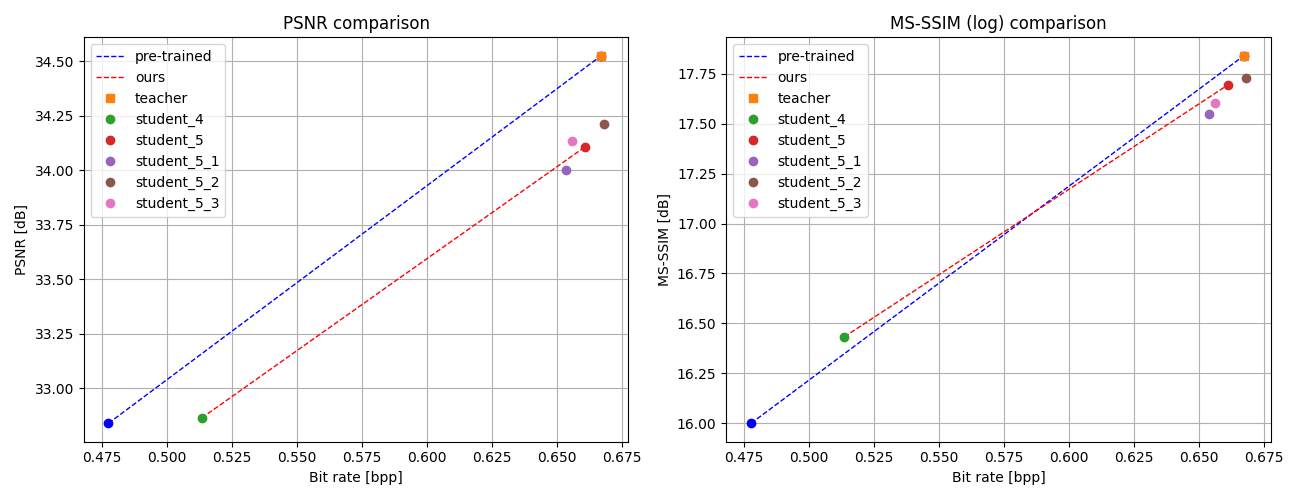}
    \caption[Average \acrshort{rd} on the Kodak dataset for students with different \acrshort{kd} tradeoffs.]{Average \acrshort{rd} on the Kodak dataset for students with different \acrshort{kd} tradeoffs. Hyper-parameters in the loss function adjust the tradeoff between \acrshort{kd} and \acrshort{rd} learning resulting in different levels of overall performance. The correct balance between \acrshort{kd} and \acrshort{rd} should be found to achieve the best results.}
    \label{kd_lic_5}
\end{figure}

Until now we used the following set of hyperparameters: \((\lambda_{1}, \lambda_{2}, \lambda_{3}) = (0.2, 0.2, 0.4)\) and varying \(\lambda\) in the \acrshort{rd} loss. We now fix \(\lambda = 0.025\) (the value used by \textsf{student\_5} in Figure \ref{kd_lic_4}) and train 3 new models with different sets of hyperparameters. Keeping in mind that these models should prioritize image quality, we can analyse Figure \ref{kd_lic_5} and focus on the distortion axis. \textsf{student\_5\_1} with \((\lambda_{1}, \lambda_{2}, \lambda_{3}) = (0.1, 0.1, 0.8)\) is the less performant due the lower weigthing of \acrshort{kd}. \((0.3, 0.3, 0.4)\) are the hyperparameters of \textsf{student\_5\_2}. Giving more weigth to the \acrshort{kd} results in better results as the student operates more like the teacher. However, giving to much weight (e.g. \((0.4, 0.4, 0.2)\)) is counter-productive. It seems that by trying to follow to closely the teacher, \textsf{student\_5\_3} does not manage as well to learn from the \acrshort{rd} loss feedback. In order to gain the most from \acrshort{kd}, the right balance should be found between \acrshort{kd} and \acrshort{rd} learning.

\begin{figure}
    \centering
    \includegraphics[width=15cm]{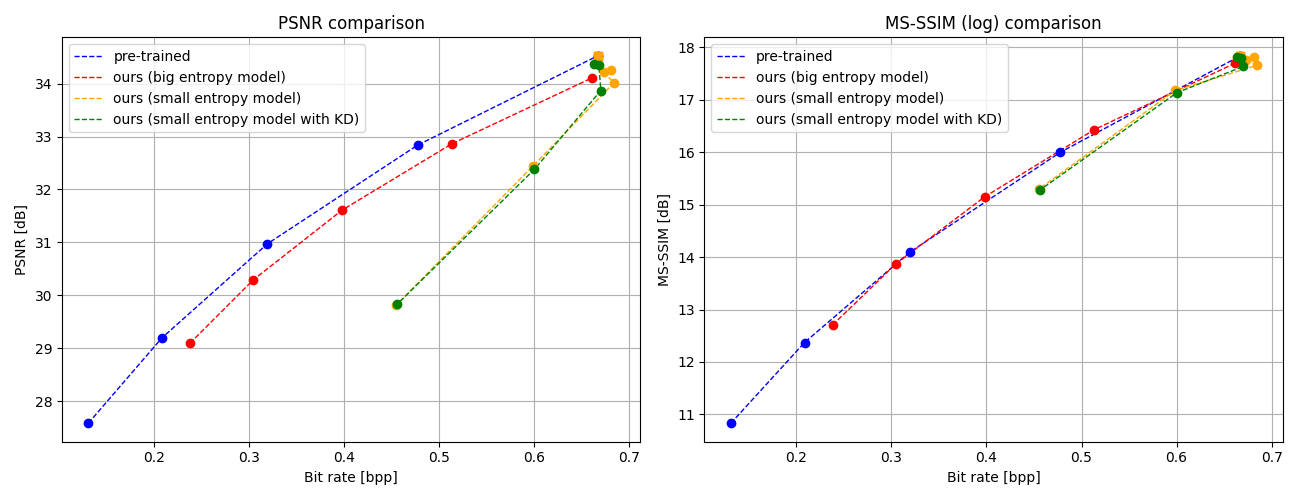}
    \caption[Average \acrshort{rd} on the Kodak dataset for students with \acrshort{kd} on the hyper-latent space.]{Average \acrshort{rd} on the Kodak dataset for students with \acrshort{kd} on the hyper-latent space. Extending \acrshort{kd} to the hyper-latent space does not bring significant overall improvements in our settings.}
    \label{kd_lic_6}
\end{figure}

\begin{figure}
    \centering
    \includegraphics[width=15cm]{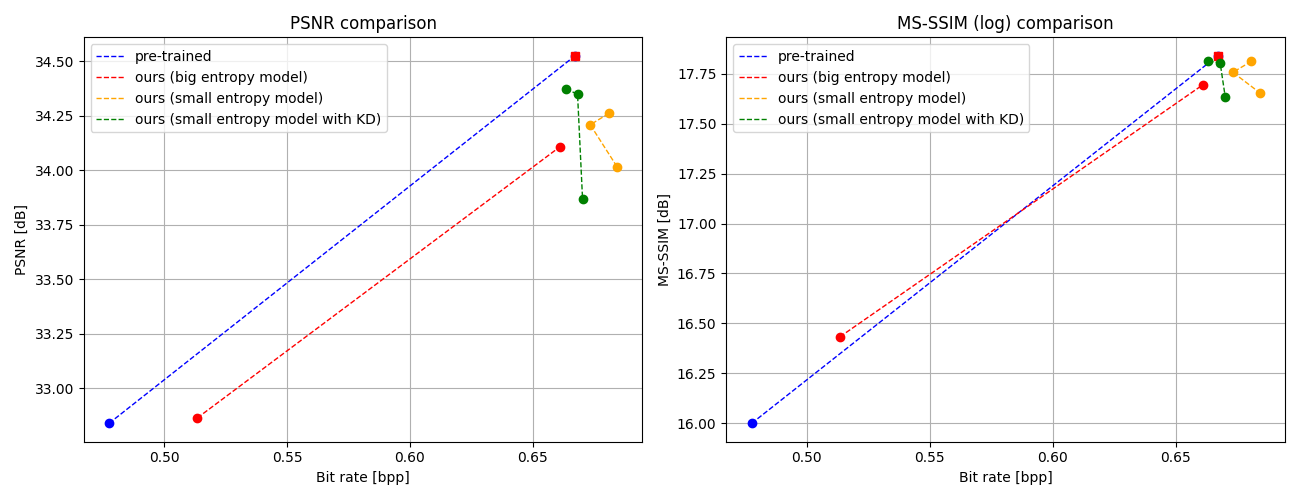}
    \caption[Average \acrshort{rd} on the Kodak dataset for students with \acrshort{kd} on the hyper-latent space (zoom).]{Average \acrshort{rd} on the Kodak dataset for students with \acrshort{kd} on the hyper-latent space. Extending \acrshort{kd} to the hyper-latent space only slightly improve results for bigger student models.}
    \label{kd_lic_7}
\end{figure}

Applying the \acrshort{kd} paradigm to the entropy model (i.e. the hyper-latent space) using Equation \ref{loss_2} shows interesting results. The students models are smaller than previous students because we also reduce the number of channels in the entropy model\footnote{The entropy model has the same number of channels as the main \acrshort{ae} apart from the last layer where it matches the number of channels of the teacher entropy model (128).}. This explains the different \acrshort{rd} performance shown in Figure \ref{kd_lic_6}. However, in our settings, the gain of applying \acrshort{kd} on the hyper-latent space only appears for models with higher number of parameters and is very small (Figure Figure \ref{kd_lic_7}). With the goal of amplifying the effect of \acrshort{kd} on the hyper-latent space, we try to train a model with two teachers. We call this method hybrid knowledge distillation. Keeping the same teacher for distilling the main \acrshort{ae}, we use a second teacher focusing on image compression (compressAI model with \textsf{quality} set to 1) on the hyper-latent space part of the loss function. The expected gain in bit rate is negligible compared to the loss in image quality (Figure \ref{kd_lic_8}). We did not conduct more experiments with these settings, as the gains seemed minor compared to the increase in resources required for training (3 GPUs). These experiments reveal other tunable hyperparameters for \acrshort{kd} training.

This section proves that \acrshort{kd} can be successfully applied to \acrshort{lic} tasks as we are able to train small models with \acrshort{rd} performance almost on par with larger models. The size of the model plays a major role in its performance. We also notice that \acrshort{kd} offers a infinite framework to tune the \acrshort{rd} performance of the student models. But \acrshort{rd} performance is only one part of the equation with frugal \acrshort{ai}: we now need to assess the resource savings. It remains to be seen whether it is worth making the trade-off of losing some \acrshort{rd} performance to use smaller models.

\subsection{Application to resource-constrained Platforms}
\label{application_resource_contrained_platforms}
Real life applications for \acrshort{lic} do not only focus on \acrshort{rd} performance. While it is important to ensure great image quality at the lowest bit rate possible, other parameters need to be taken into account. Having proved the effectiveness of \acrshort{kd} for \acrshort{lic} tasks in terms of \acrshort{rd} performance, we now need to assess its benefits in the context of resource-limited platforms. This section analyses student models with different architectures from a resource stand point across three main axis: memory, computing power, and energy consumption.

In this section, we use the models first introduced in Section \ref{architecture_size_tradeoff}. There are five student architectures with 16, 32, 64, 96 and 112 channels and a latent space of size 192. We compare our models to pre-trained models from the compressAI zoo with \textsf{quality} ranging from 1 to 5.

\begin{figure}
    \centering
    \includegraphics[width=15cm]{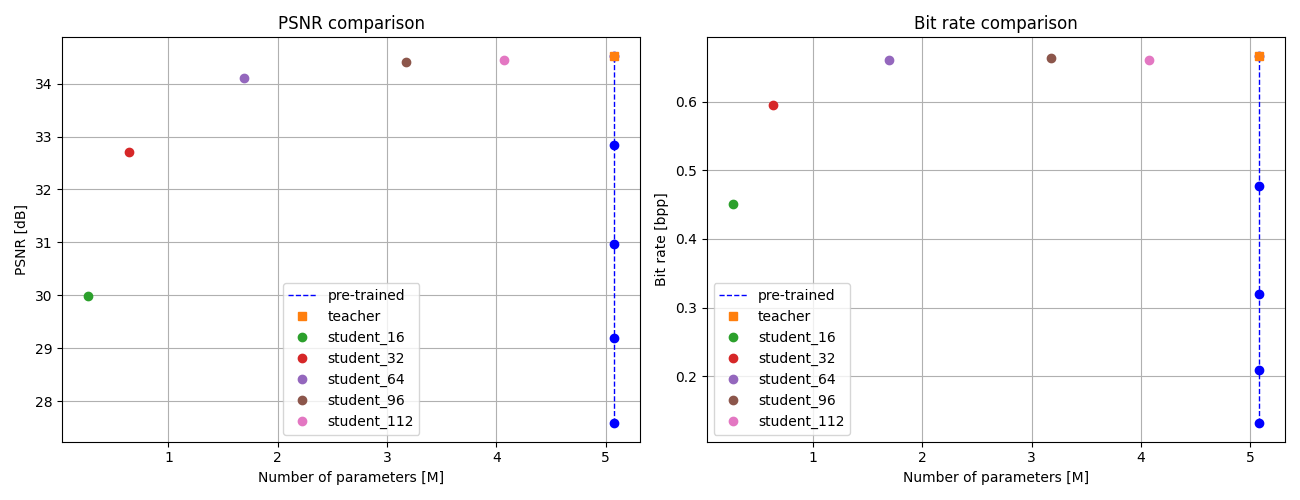}
    \caption[\acrshort{psnr} and bit rate on the Kodak dataset according to students number of parameters.]{\acrshort{psnr} and bit rate on the Kodak dataset according to students number of parameters. Reducing the number of parameters gradually decrease \acrshort{rd} results. Similar results are observed for memory footprint. In our testing, a fair balance between \acrshort{rd} and memory footprint is using 64 channels as it virtually does not degrade performance while reducing the memory footprint by 68 \%.}
    \label{kd_lic_parameters}
\end{figure}

Most resource-constrained computers deal with a limited amount of memory. This limiting factor sometimes makes them unsuitable for tasks requiring large models. In \acrshort{lic}, distillation allows the use of smaller models without degrading \acrshort{rd} performance. Our student models can have as few as 0.27 M parameters while standard models have 5 M parameters. This is a reduction of up to 95 \% in terms of parameters or memory size. The values for each student model are presented in Table \ref{tab_size}. Although the number of parameters definitely has an impact on \acrshort{rd} capabilities of \acrshort{nn}, Figures \ref{kd_lic_parameters} and \ref{appendix:kd_lic_memory} show that student models with 64 channels and more are quite close to their teacher in both \acrshort{psnr} and bit rate. In other words, these models could be used instead of the teacher on devices with limited memory without noticeably degrading the user experience. The student with 64 channels is particularly valuable, as it offers \acrshort{psnr} and bit rate inline with the teacher while reducing the memory footprint by 68 \%.

\sisetup{
  detect-weight=true,
  detect-inline-weight=math,
  table-number-alignment=center,
  table-figures-integer=3,
  table-figures-decimal=2
}

\begin{table}[t]
    \centering
    \caption{Number of parameters, memory footprint, \acrshort{psnr}, and bit rate for teacher and student models.}
    \label{tab_size}
    \begin{tabular}{l c
                    S[table-format=1.2] S[table-format=2.2]
                    S[table-format=2.2] S[table-format=2.2]
                    S[table-format=2.2] S[table-format=2.2]
                    S[table-format=1.2] S[table-format=1.2]}
        \toprule
        Model & Channels &
        \multicolumn{2}{c}{Parameters [M]} &
        \multicolumn{2}{c}{Memory [MB]} &
        \multicolumn{2}{c}{PSNR} &
        \multicolumn{2}{c}{Bit rate [bpp]} \\
        \cmidrule(lr){3-4} \cmidrule(lr){5-6} \cmidrule(lr){7-8} \cmidrule(lr){9-10}
        & & {Abs.} & {\%$\Delta$} & {Abs.} & {\%$\Delta$} & {Abs.} & {\%$\Delta$} & {Abs.} & {\%$\Delta$} \\

        Teacher & 128 & 5.08 & { }      & 20.18 & { }      & 34.53 & { }       & 0.67 & { } \\
        Student & 112 & 4.07 & \textcolor{gray}{-19.8} & 15.53 & \textcolor{gray}{-23.0} & 34.44 & \textcolor{gray}{-0.3} & 0.66 & \textcolor{gray}{-1.0} \\
                & 96  & 3.17 & \textcolor{gray}{-37.5} & 12.11 & \textcolor{gray}{-40.0} & 34.41 & \textcolor{gray}{-0.3} & 0.66 & \textcolor{gray}{-0.6} \\
                & 64  & 1.69 & \textcolor{gray}{-66.6} & 6.46  & \textcolor{gray}{-68.0} & 34.11 & \textcolor{gray}{-1.2} & 0.66 & \textcolor{gray}{-0.9} \\
                & 32  & 0.64 & \textcolor{gray}{-87.5} & 2.43  & \textcolor{gray}{-88.0} & 32.71 & \textcolor{gray}{-5.3} & 0.60 & \textcolor{gray}{-10.7} \\
                & 16  & 0.27 & \textcolor{gray}{-94.8} & 1.01  & \textcolor{gray}{-95.0} & 29.98 & \textcolor{gray}{-13.2} & 0.45 & \textcolor{gray}{-32.4} \\
        \bottomrule
    \end{tabular}
\end{table}

\begin{figure}
    \centering
    \includegraphics[width=15cm]{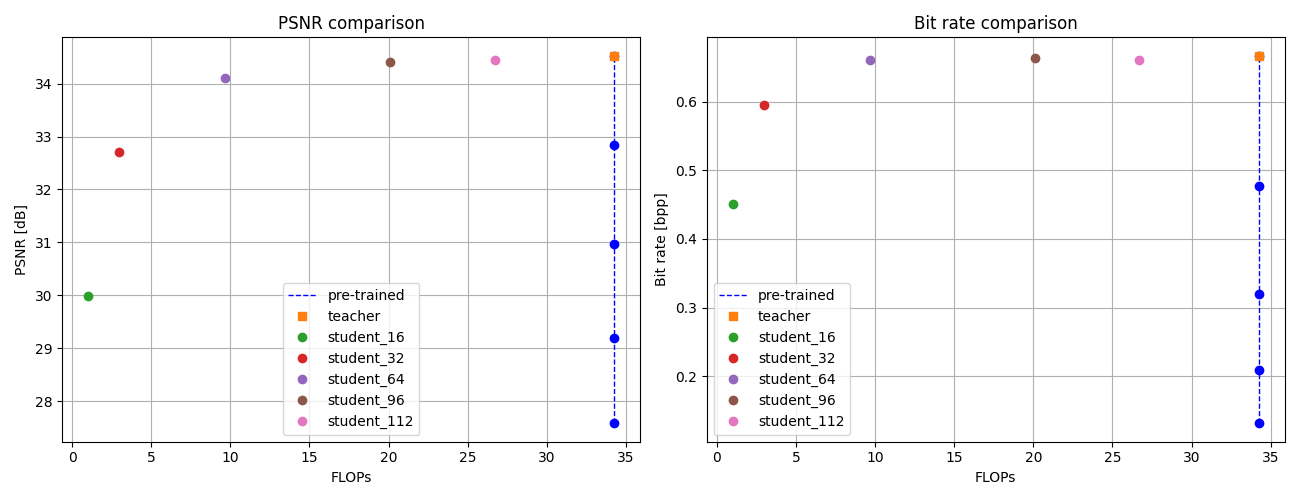}
    \caption[\acrshort{psnr} and bit rate on the Kodak dataset according to students \acrshort{flop}s.]{\acrshort{psnr} and bit rate on the Kodak dataset according to students \acrshort{flop}s. Our knowledge distilled models are good candidates for devices with limited compute power. They all offer lower \acrshort{flop} counts than standard models. However, this comes at a non-negligible cost in image quality and bit rate for the smallest models.}
    \label{kd_lic_flop}
\end{figure}

\begin{figure}
    \centering
    \includegraphics[width=15cm]{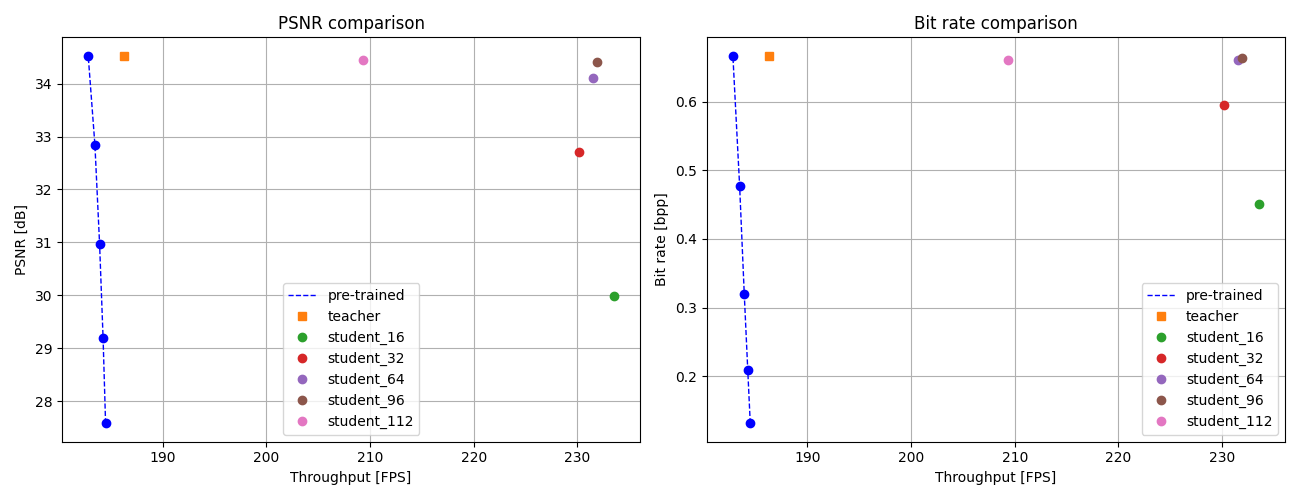}
    \caption[\acrshort{psnr} and bit rate on the Kodak dataset according to students' throughput.]{\acrshort{psnr} and bit rate on the Kodak dataset according to students throughput. All models, exceed standard frame rate. With, higher throughput than pre-trained models, our knowledge distilled models have more headroom for higher resolution frames or lower energy consumption depending the application.}
    \label{kd_lic_fps}
\end{figure}

Computers can only perform a certain amount of operations per unit of time. When using with mainstream hardware, the computing power required to use an image compression model like the scale hyperprior model is sufficient. However, when dealing with resource-constrained devices, the latency might increase which goes against the objective of real-time decompression. With too much latency, it is impossible to extend image decompression to video decompression. Here, we focus on two metrics: the number of \acrfull{flop} carried out to run an inference and the model throughput (i.e. the number of frames processed by the model in one second). According to our results, the student with 16 channels only requires 3 \% of the teacher \acrshort{flop}s to perform the inference which translates to a 25 \% increase in throughput (Table \ref{tab_compute}). Figures \ref{kd_lic_flop} and \ref{kd_lic_fps} show that once again, the student with 64 channels represents the best compromise between \acrshort{rd} and \acrshort{flop}s or throughput. It should also be noted that all our models present a throughput that exceeds all requirements for video streaming. This headroom can be used in different ways: we can either increase the stream resolution to enhance user experience or reduce the inference frequency to save energy. 

\sisetup{
  detect-weight=true,
  detect-inline-weight=math,
  table-number-alignment=center,
  table-figures-integer=3,
  table-figures-decimal=2
}

\begin{table}[t]
    \centering
    \caption{\acrshort{flop}s, throughput, \acrshort{psnr}, and bit rate for teacher and student models.}
    \label{tab_compute}
    \begin{tabular}{l c
                    S[table-format=2.2] S[table-format=2.1]
                    S[table-format=3.2] S[table-format=3.2]
                    S[table-format=2.2] S[table-format=2.2]
                    S[table-format=1.2] S[table-format=2.2]}
        \toprule
        Model & Channels &
        \multicolumn{2}{c}{GFLOPs/frame} &
        \multicolumn{2}{c}{Throughput [FPS]} &
        \multicolumn{2}{c}{PSNR} &
        \multicolumn{2}{c}{Bit rate [bpp]} \\
        \cmidrule(lr){3-4} \cmidrule(lr){5-6} \cmidrule(lr){7-8} \cmidrule(lr){9-10}
        & & {Abs.} & {\%$\Delta$} & {Abs.} & {\%$\Delta$} & {Abs.} & {\%$\Delta$} & {Abs.} & {\%$\Delta$} \\
        
        Teacher & 128 & 34.24 & { } & 184.20 & { } & 34.53 & { } & 0.67 & { } \\
        Student & 112 & 26.70 & \textcolor{gray}{-22.0} & 209.01 & \textcolor{gray}{+13.5} & 34.44 & \textcolor{gray}{-0.3} & 0.66 & \textcolor{gray}{-1.0} \\
                & 96  & 20.10 & \textcolor{gray}{-41.3} & 231.61 & \textcolor{gray}{+25.7} & 34.41 & \textcolor{gray}{-0.3} & 0.66 & \textcolor{gray}{-0.6} \\
                & 64  & 9.67  & \textcolor{gray}{-71.8} & 232.47 & \textcolor{gray}{+26.2} & 34.11 & \textcolor{gray}{-1.2} & 0.66 & \textcolor{gray}{-0.9} \\
                & 32  & 2.98  & \textcolor{gray}{-91.3} & 231.90 & \textcolor{gray}{+25.9} & 32.71 & \textcolor{gray}{-5.3} & 0.60 & \textcolor{gray}{-10.7} \\
                & 16  & 1.02  & \textcolor{gray}{-97.0} & 233.63 & \textcolor{gray}{+26.8} & 29.98 & \textcolor{gray}{-13.2} & 0.45 & \textcolor{gray}{-32.4} \\
        \bottomrule
    \end{tabular}
\end{table}

\begin{figure}
    \centering
    \includegraphics[width=15cm]{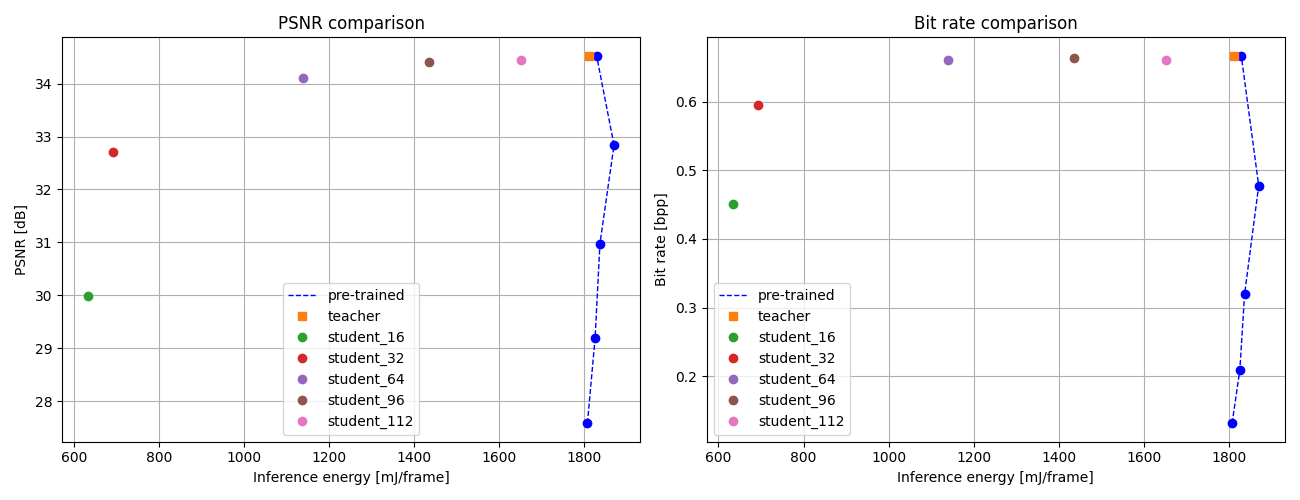}
    \caption[\acrshort{psnr} and bit rate on the Kodak dataset according to students consumed energy.]{\acrshort{psnr} and bit rate on the Kodak dataset according to students consumed energy. \acrshort{kd} is a great method to create energy friendly \acrshort{lic} models. If energy consumption is not too restricted, we recommend using the student with 64 channels that consumes 35 \% less energy per frame and achieve image compression on par with the teacher model.}
    \label{kd_lic_energy}
\end{figure}

Most edge devices also have access to a limited amount of energy whether it is in time because they run on battery like smartphones or because they are \acrfull{iot} devices that run 24/7 and thus should not consume a lot of energy. This is why we focus on the energy required to process a single frame. Table \ref{tab_energy} shows that we can save up to 60 \% of the energy used by the teacher model by using the student with 16 channels. Using Figure \ref{kd_lic_energy}, that the model that offers the best tradeoff between \acrshort{psnr} and energy consumption in the student model with 64 channels. By using this model we keep the same image quality while reducing our energy consumption by 35 \%.

\sisetup{
  detect-weight=true,
  detect-inline-weight=math,
  table-number-alignment=center,
  table-figures-integer=4,
  table-figures-decimal=2
}

% Energy table
\begin{table}[t]
    \centering
    \caption{Consumed energy, \acrshort{psnr}, and bit rate for teacher and student models.}
    \label{tab_energy}
    % wider layout for this table only
    \setlength{\tabcolsep}{12pt}
    \renewcommand{\arraystretch}{1.2}
    \begin{tabular}{l c
                    S[table-format=4.2] S[table-format=2.2]
                    S[table-format=2.2] S[table-format=2.2]
                    S[table-format=1.2] S[table-format=2.2]}
        \toprule
        Model & Channels &
        \multicolumn{2}{c}{Energy [mJ/frame]} &
        \multicolumn{2}{c}{PSNR} &
        \multicolumn{2}{c}{Bit rate [bpp]} \\
        \cmidrule(lr){3-4} \cmidrule(lr){5-6} \cmidrule(lr){7-8}
        & & {Abs.} & {\%\(\Delta\)} & {Abs.} & {\%\(\Delta\)} & {Abs.} & {\%\(\Delta\)} \\
        \midrule
        Teacher & 128 & 1767.85 & { } & 34.53 & { } & 0.67 & { } \\
        Student & 112 & 1600.41 & \textcolor{gray}{-9.5}  & 34.44 & \textcolor{gray}{-0.3}  & 0.66 & \textcolor{gray}{-1.0} \\
                & 96  & 1416.34 & \textcolor{gray}{-19.9} & 34.41 & \textcolor{gray}{-0.3}  & 0.66 & \textcolor{gray}{-0.6} \\
                & 64  & 1164.17 & \textcolor{gray}{-34.2} & 34.11 & \textcolor{gray}{-1.2}  & 0.66 & \textcolor{gray}{-0.9} \\
                & 32  & 699.18  & \textcolor{gray}{-60.5} & 32.71 & \textcolor{gray}{-5.3}  & 0.60 & \textcolor{gray}{-10.7} \\
                & 16  & 684.45  & \textcolor{gray}{-61.3} & 29.98 & \textcolor{gray}{-13.2} & 0.45 & \textcolor{gray}{-32.4} \\
        \bottomrule
    \end{tabular}
\end{table}

% Codec table
\begin{table}[t]
    \centering
    \caption{Throughput, consumed energy, \acrshort{psnr}, and bit rate for different codecs.}
    \label{tab_codecs}
    \setlength{\tabcolsep}{14pt}
    \renewcommand{\arraystretch}{1.2}
    \begin{tabular}{lcccc}
        \toprule
        Codec & Throughput [FPS] & Energy [mJ/frame] & PSNR & Bit rate [bpp] \\
        \midrule
        JPEG 2000 & 10.15 & 2308.87 & 70.0 & 13.45 \\
        \midrule
        \multirow{3}{*}{JPEG}
          & 130.12 & \cellcolor{Good}185.84 & 26.65 & 0.33 \\
          & \cellcolor{Good}134.30 & \cellcolor{Bad}804.04 & \cellcolor{Bad}21.41 & \cellcolor{Good}0.17 \\
          & \cellcolor{Bad}86.51 & 274.94 & \cellcolor{Good}40.55 & \cellcolor{Bad}3.40 \\
        \midrule
        \multirow{5}{*}{WebP}
          & \cellcolor{Bad}19.11 & 1216.75 & 43.03 & 3.85 \\
          & \cellcolor{Good}39.61 & 603.38 & 25.85 & 0.11 \\
          & 39.56 & \cellcolor{Good}595.61 & \cellcolor{Bad}25.85 & \cellcolor{Good}0.11 \\
          & 19.12 & 1220.86 & \cellcolor{Good}43.03 & \cellcolor{Bad}3.85 \\
          & 19.14 & \cellcolor{Bad}1232.84 & 43.03 & 3.85 \\
        \bottomrule
    \end{tabular}
\end{table}

In order to put our results into context, we also measured throughput, energy consumption, \acrshort{psnr} and bit rate for three image compression codecs, namely: JPEG-2000, JPEG and WepP. The first row of Table \ref{tab_energy} shows that JPEG-2000 is not suited for our application due to its emphasis on image quality: the bit rate and energy consumption are to high. In addition to this, the throughput is very low making it unusable in real-time. Our experiments show that studied \acrshort{lic} models outperform JPEG and WepP in \acrshort{rd} (Figure \ref{appendix:codecs_rd}). Both JPEG and WepP could be used in real-time albeit at a lower frame rate than our \acrshort{nn}s (Figure \ref{appendix:codecs_fps}). However, when it comes to energy consumption, the superiority of \acrshort{lic} models is challenged: JPEG achieves better \acrshort{psnr} and bit rate with less energy (Figure \ref{appendix:codecs_energy}). This brief comparison with traditional codecs show that \acrshort{lic} models, and especially our distilled models, are relevant candidates for real-time image compression on resource-constrained platforms. These models propose an interesting compromise between throughput, energy consumption and quality of image quality.

This section is a deep dive into the resource consumption of standard and knowledge distilled models. Based on measures of memory usage, required computing power and energy consumption on students architectures with varying number of channels, we prove that \acrshort{kd} indeed create frugal models. In all experiments, our \acrshort{kd} models consume less resources that standard models at a (sometimes negligible) cost in \acrshort{rd} performance. On a not-too-constrained platform, we would recommend the use of the student with 64 channels. It has a limited use of resources without noticeable impact on image compression.

This section is a step in the right direction for \acrshort{lic} on resource-constrained platforms. We first assess the effectiveness of \acrshort{kd} on \acrshort{lic} using state-of-the-art models. \acrshort{rd} results are impressive, showing models with as low as half as many channels in the same operational region as their teacher. We also experiment with different teachers and find that we can train student models adapted for any image compression/resource consumption tradeoff possible. Other settings like the loss function definitely impact the student performances, leaving room for other research work. Most importantly, we measure a significant reduction in the model resource consumption (memory, compute, and energy) without too much degradation of the image compression \acrshort{rd}. We observe that \acrshort{kd} is a valuable training paradigm in order to achieve real-time image compression on resource limited devices.

%#### Conclusion ##############################################################
\section{Conclusion}
Data compression is critical in information technology, helping save storage space and improve network efficiency by reducing the amount of bits needed to represent information. While lossless compression has its limitations, lossy compression introduces a tradeoff between storage efficiency and data quality. Deep learning has opened new possibilities for image compression, but it often requires significant computational power, which can limit its use on resource-constrained devices.

This research aims to explore methods for enabling image compression with deep learning models on low-power devices, potentially in real-time. We focused on \acrfull{kd}, a technique that trains smaller, more efficient models (students) with guidance from larger, high-performing models (teachers). Our experiments showed that KD enhances performance in image compression tasks, achieving results comparable to teacher models while significantly reducing resource consumption. We observe a reduction in memory usage, processing power, and energy consumption without significantly affecting compression performance. Our results highlight the importance of selecting the right tradeoff between image quality and resource efficiency, depending on the application. We explored different hyperparameters for the training of students such as: \acrfull{nn} size, different weight parameters in the loss function, \acrshort{kd} on the hyper-latent space and \acrshort{kd} with multiple teachers. 

Looking ahead, we plan to explore hybrid architectures in which larger encoders can be used with smaller decoders. This could offer further improvements, especially in scenarios where powerful servers compress images for broadcasting to multiple edge devices.

However, challenges remain, particularly in the complexity of the KD training process, which requires more resources and depends on various hyperparameters and teacher quality. Future research will also need to consider data dependencies and the transparency of models, especially in contexts where model explainability is critical. \acrlong{kd}, by focusing on output alone, can result in less interpretability compared to traditional models, raising concerns in regulated environments.

%#### Bibliography ############################################################
\printbibliography

%#### Appendix ################################################################
\appendix

\section{Reproducibility}

\subsection{Implementation Details}
All experiments are conducted using Python 3.12.7 and the version 1.2.6 of compressAI (see requirements file for other library version). We use an Anaconda virtual environment on the Télécom Paris GPU cluster which provides the processing power to perform our trainings and testings.

Measures of Section \ref{application_resource_contrained_platforms} are obtained using PyTorch properties. By iterating through \textsf{model.parameters()}, it is possible to compute the number of parameters and the memory footprint of a model.

\acrshort{flop}s are measured once per model using the \textsf{FlopCountAnalysis} from the fvcore library. We use two libraries to compute the inference time and energy consumption of the models: pynvml and zeus. By iterating 50 times through the test dataset images loaded on the same Nvidia RTX 3090 GPU beforehand, we are able to achieve meaningful average values per frame. Both libraries present similar results and we ultimately chose the zeus library for our final results. The value of inference time per frame is used to compute the model throughput.

In Section \ref{application_resource_contrained_platforms}, we use the PIL Python library to convert images from the Kodak dataset to JPEG-2000, JPEG and WepP. We follow the same method as for evaluating \acrshort{nn}s.

\section{Figures}

\begin{figure}
  \centering
  \includegraphics[width=15cm]{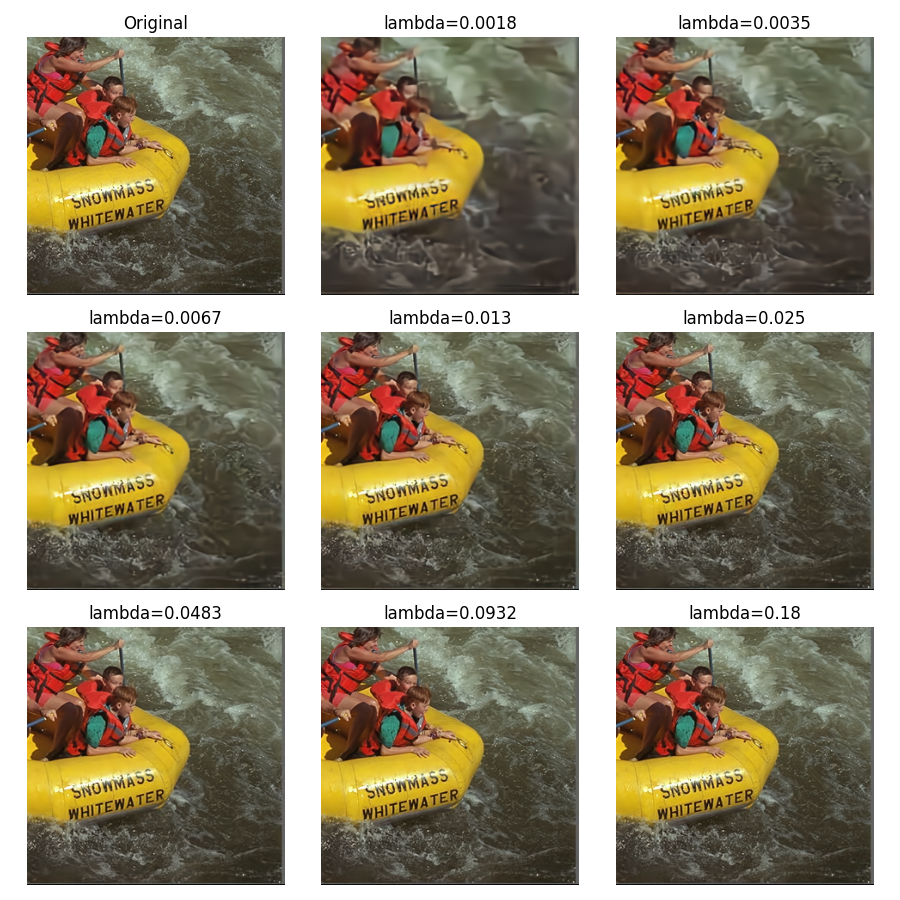}
  \caption[Reconstruction results on image 14 of the Kodak dataset for different RD tradeoffs (pre-trained models).]{Reconstruction results on image 14 of the Kodak dataset for different RD tradeoffs (pre-trained models).}
  \label{appendix:pretrained}
\end{figure}

\begin{figure}
  \centering
  \includegraphics[width=15cm]{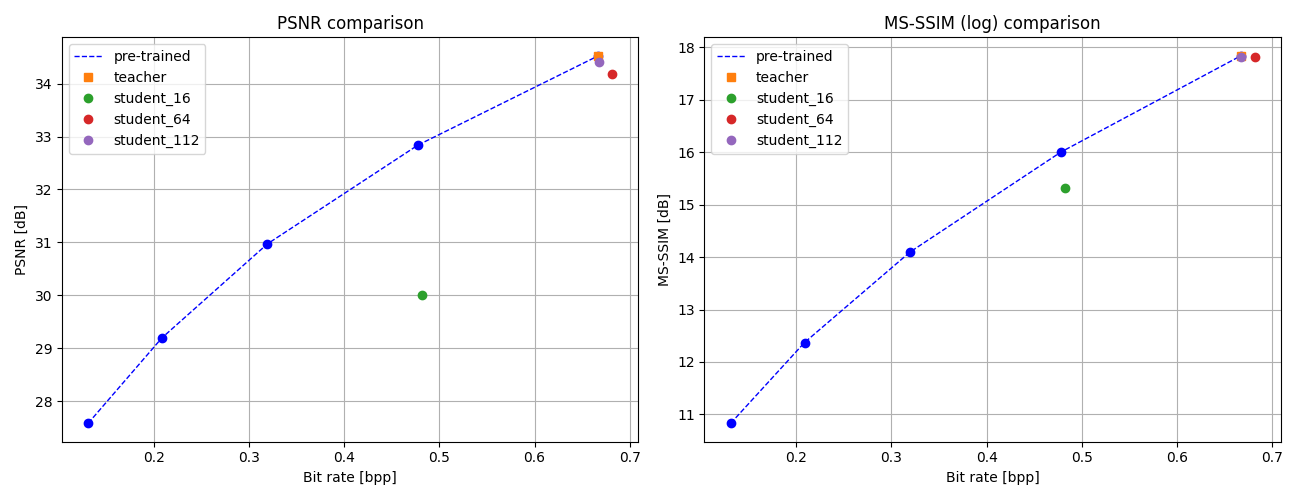}
  \caption[Average \acrshort{rd} curve on the Kodak dataset for students with different number of channels and Kullback-Leibler divergence loss on the latent space.]{Average \acrshort{rd} curve on the Kodak dataset for students with different number of channels and Kullback-Leibler divergence loss on the latent space.}
  \label{appendix:kd_lic_1_kld}
\end{figure}

\begin{figure}
  \centering
  \includegraphics[width=15cm]{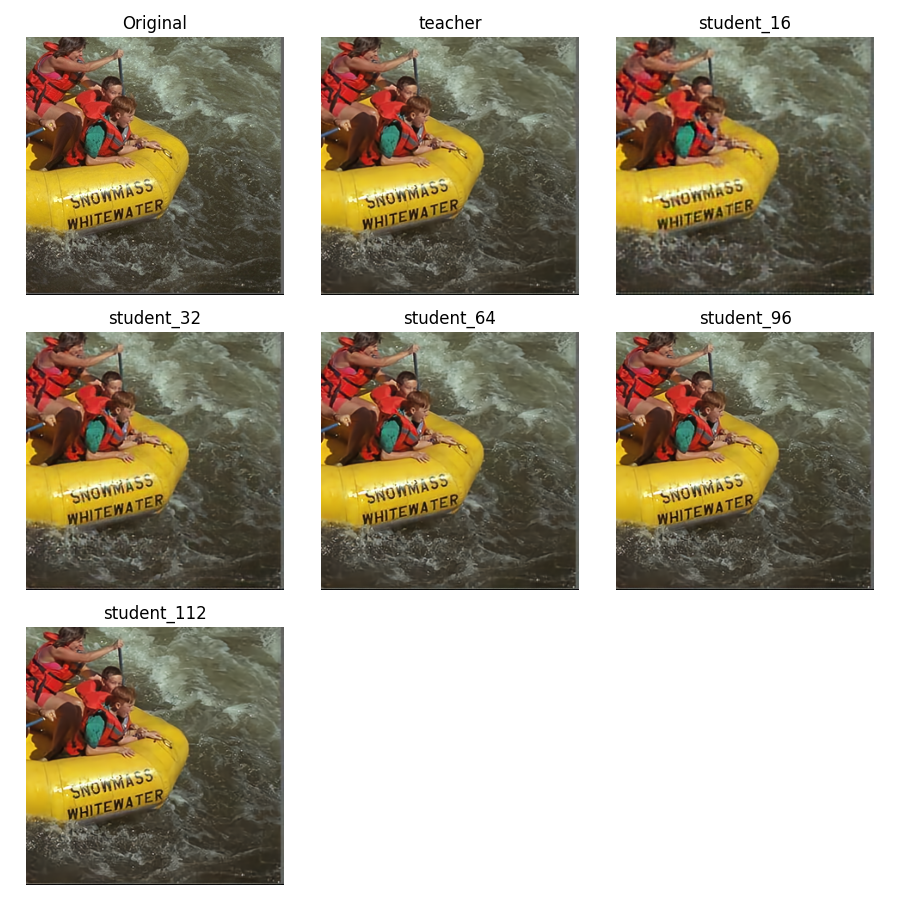}
  \caption[Reconstruction results on image 14 of the Kodak dataset for image compression with different architectures.]{Reconstruction results on image 14 of the Kodak dataset for image compression with different architectures (pre-trained teacher with \textsf{quality} set to 5 and our student models with different number of channels trained for image compression, see Section \ref{architecture_size_tradeoff}).}
  \label{appendix:kd_lic_2:a}
\end{figure}

\begin{figure}
    \centering
    \includegraphics[width=15cm]{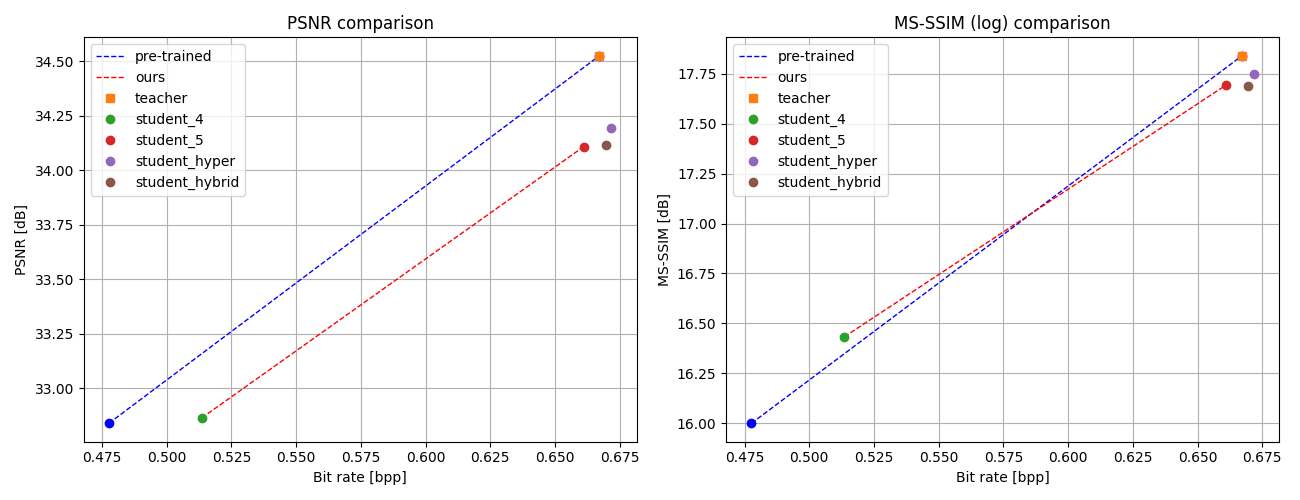}
    \caption[Average \acrshort{rd} on the Kodak dataset for student with hybrid \acrshort{kd}.]{Average \acrshort{rd} on the Kodak dataset for student with hybrid \acrshort{kd}.}
    \label{kd_lic_8}
\end{figure}

\begin{figure}
  \centering
  \includegraphics[width=15cm]{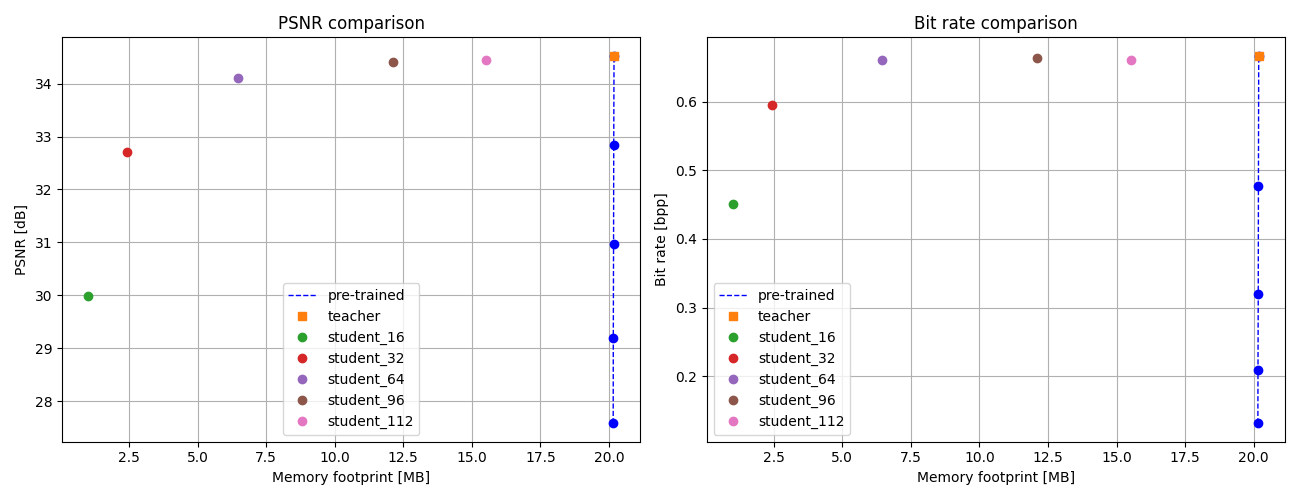}
  \caption[\acrshort{psnr} and bit rate on the Kodak dataset according to students memory footprint.]{\acrshort{psnr} and bit rate on the Kodak dataset according to students memory footprint.}
  \label{appendix:kd_lic_memory}
\end{figure}

\begin{figure}
  \centering
  \includegraphics[width=15cm]{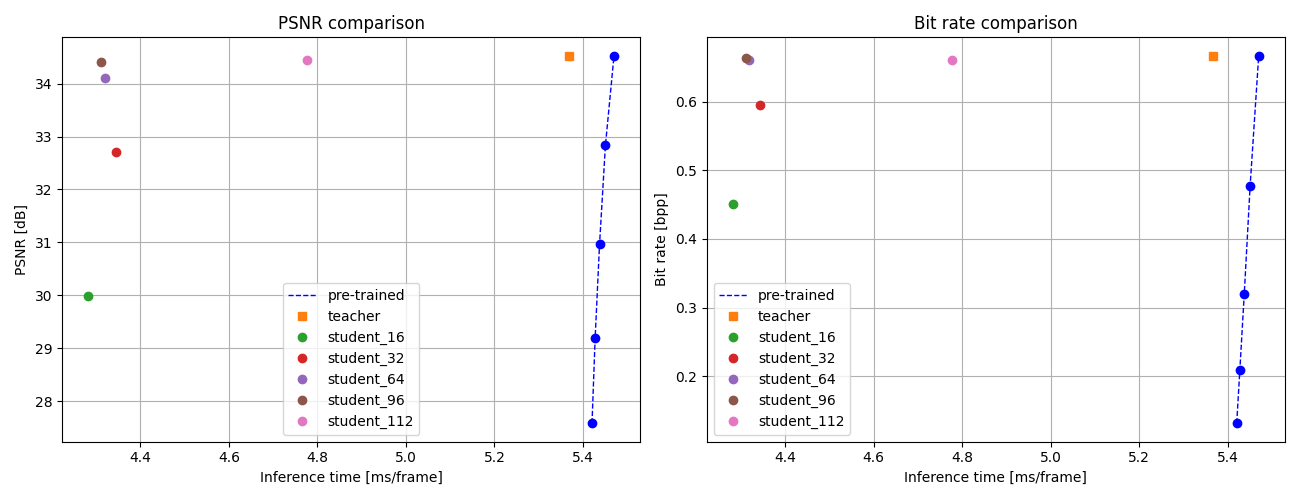}
  \caption[\acrshort{psnr} and bit rate on the Kodak dataset according to students inference time.]{\acrshort{psnr} and bit rate on the Kodak dataset according to students inference time.}
  \label{appendix:kd_lic_time}
\end{figure}

\begin{figure}
  \centering
  \includegraphics[width=15cm]{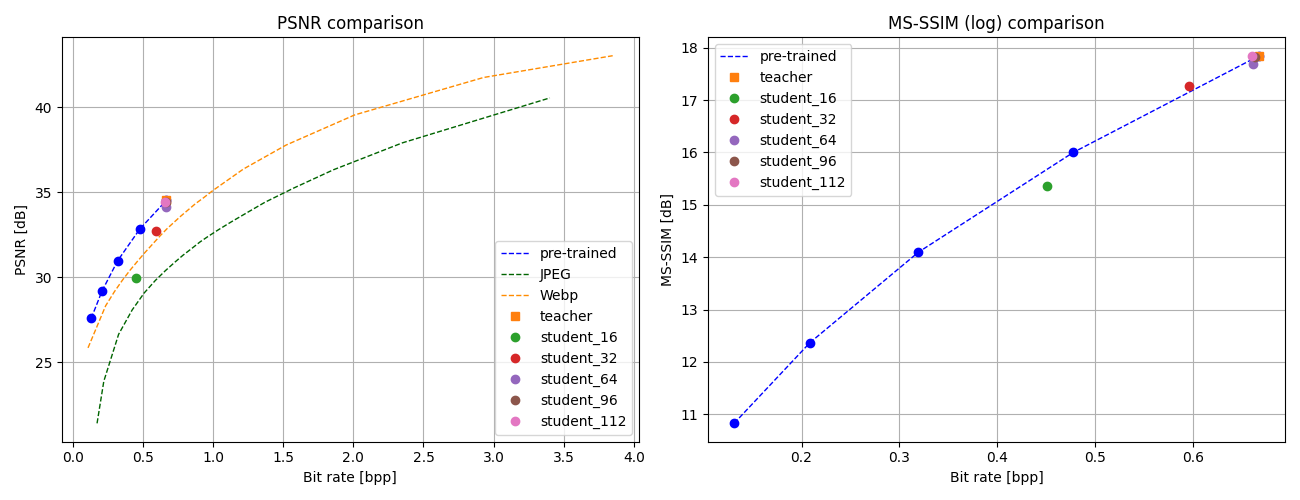}
  \caption[Average \acrshort{rd} curve on the Kodak dataset for students with different number of channels and codecs.]{Average \acrshort{rd} curve on the Kodak dataset for students with different number of channels and codecs.}
  \label{appendix:codecs_rd}
\end{figure}

\begin{figure}
  \centering
  \includegraphics[width=15cm]{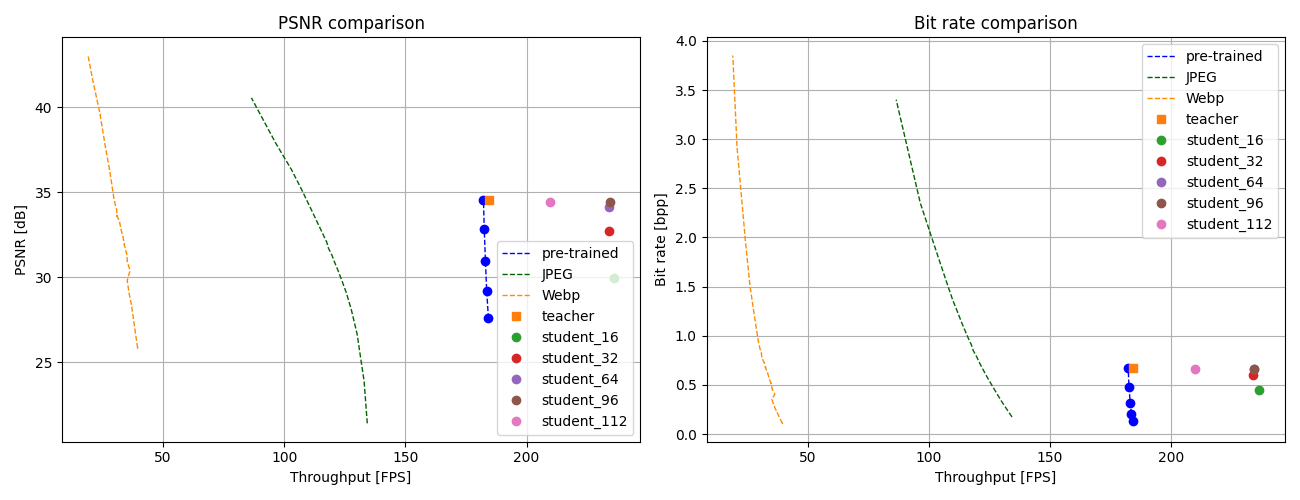}
  \caption[\acrshort{psnr} and bit rate on the Kodak dataset according to students and codecs throughput.]{\acrshort{psnr} and bit rate on the Kodak dataset according to students and codecs throughput.}
  \label{appendix:codecs_fps}
\end{figure}

\begin{figure}
  \centering
  \includegraphics[width=15cm]{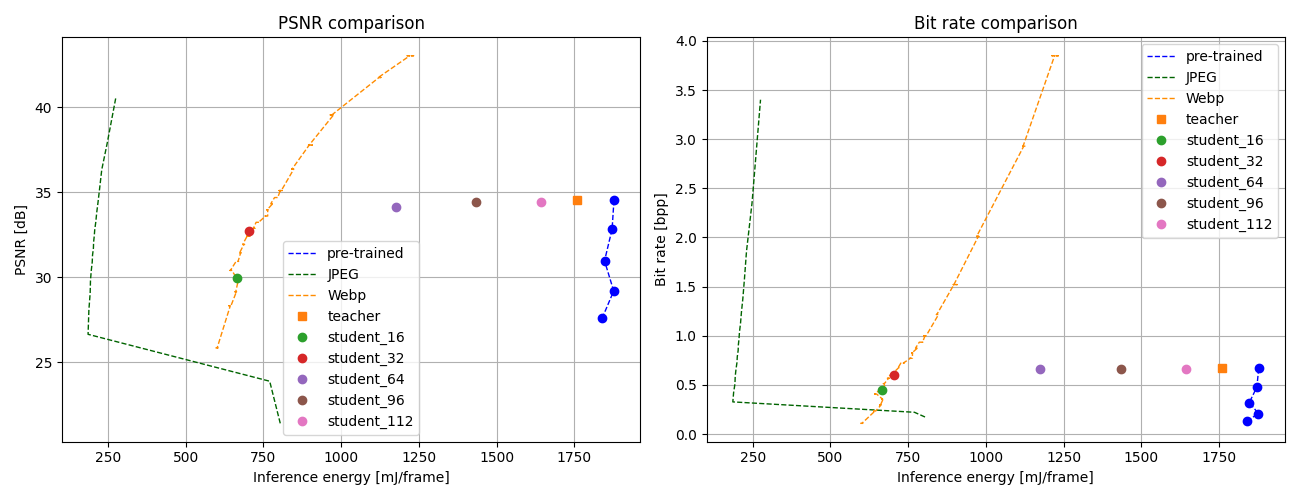}
  \caption[\acrshort{psnr} and bit rate on the Kodak dataset according to students and codecs consumed energy.]{\acrshort{psnr} and bit rate on the Kodak dataset according to students and codecs consumed energy.}
  \label{appendix:codecs_energy}
\end{figure}

\end{document}